\documentclass[10pt,journal,compsoc]{IEEEtran}
\usepackage{amsfonts}
\usepackage{amsmath}
\usepackage{amsthm}
\usepackage{wrapfig}
\usepackage[colorlinks=True,
            linkcolor=blue,
            anchorcolor=green,
            citecolor=blue
            ]{hyperref} 

\usepackage{float}
\usepackage{url}
\usepackage{graphicx}
\usepackage{subcaption} 
\usepackage{multirow}
\usepackage{enumitem}
\usepackage[ruled,vlined]{algorithm2e}

\def\BB{{\mathbb B}}
\def\D{{\mathcal D}}
\def\L{{\mathcal L}}

\def\W{{\textbf  W}}
\def\V{{\textbf  V}}
\def\Y{{\textbf  Y}}
\def\N{{\mathbb  N}}
\def\M{{\mathbb  M}}
\def\S{{\mathbb  S}}

\def\bw{\textbf{w}}
\def\bz{\textbf{z}}
\def\bx{\textbf{x}} 
\def\by{\textbf{y}}
\def\bv{\textbf{v}}

\def\bpi{{\boldsymbol  \pi}} 
\def\LA{{\boldsymbol \Lambda}}
\def\argmin{{\rm argmin}}

\def\eqspace{\arraycolsep=1.5pt\def\arraystretch}

\makeatletter
\patchcmd{\@algocf@start}
  {-1.5em}
  {0pt}
  {}{}
\makeatother
\usepackage{xpatch}

\makeatletter
\xpatchcmd\SetKwInOut
  {\hangafter=1\parbox[t]}
  {\hangafter=1\justify\parbox[t]}
  {}{\fail}
\makeatother
\usepackage{ragged2e}

\newtheorem{theorem}{Theorem}[section]

\newtheorem{lemma}{Lemma}[section]

\newtheorem{assumption}{Assumption}[section]

\begin{document}
\title{BADM: Batch ADMM for Deep Learning }


\author{Ouya Wang,
        Shenglong Zhou,
        and~Geoffrey Ye Li,~\IEEEmembership{Fellow,~IEEE}
\IEEEcompsocitemizethanks{
\IEEEcompsocthanksitem O. Wang and G. Li are with the ITP Lab, Department of Electrical and Electronic Engineering, Imperial College London, the United Kingdom ( ouya.wang20@imperial.ac.uk, geoffrey.li@imperial.ac.uk).\protect
\IEEEcompsocthanksitem S. Zhou is with the School of Mathematics and Statistics, Beijing
Jiaotong University, Beijing, China (shlzhou@bjtu.edu.cn).\protect
\IEEEcompsocthanksitem This work was supported by the Fundamental Research Funds for the Central Universities and the Talent Fund of Beijing Jiaotong University.
\IEEEcompsocthanksitem Corresponding author: Shenglong Zhou.
}

\thanks{Manuscript received April 19, 2005; revised August 26, 2015.}}

\markboth{
}%
{Shell \MakeLowercase{\textit{et al.}}: Bare Advanced Demo of IEEEtran.cls for IEEE Computer Society Journals}

\IEEEtitleabstractindextext{%
\begin{abstract}
\justifying 
Stochastic gradient descent (SGD) algorithms are widely used for training deep neural networks but often struggle with slow convergence. To address this issue, we leverage the framework of the alternating direction method of multipliers (ADMM) and develop a novel data-driven algorithm called batch ADMM (BADM).  The key innovation of BADM lies in its data-splitting strategy: the training data is divided into batches, which are further split into sub-batches. Within this structure, global parameters are aggregated in each batch, and primal and dual variables are iteratively updated using sub-batch data. We prove that BADM achieves a sublinear convergence rate under relatively mild assumptions and evaluate its performance across diverse deep learning tasks, including graph modeling, computer vision, image generation, and natural language processing. Extensive numerical experiments demonstrate that BADM achieves faster convergence and superior testing accuracy compared to other state-of-the-art optimizers. 
\end{abstract}

}
\maketitle
\newcommand{\fix}{\marginpar{FIX}}
\newcommand{\new}{\marginpar{NEW}}




\section{Introduction}

\IEEEPARstart{D}{eep} learning (DL) has revolutionized a variety of applications such as computer vision, natural language processing (NLP), image generation \cite{ho2020denoising,goodfellow2020generative},   wireless communications \cite{wang2022learn, wang2023effective}, energy systems \cite{yang2022resilient, yang2024comparative, yang2023optimising}, to name a few. Central to the success of DL models is the optimization of their parameters, which involves finding the optimal set of weights that minimize a given loss function. 

\subsection{SGD-based learning algorithms} One of the most popular and effective optimization methods for training deep neural networks (DNNs) is stochastic gradient descent (SGD)-based algorithms.  However, these algorithms frequently suffer from slow convergence, especially in high-dimensional and non-convex landscapes \cite{chen2023symbolic}, resulting in enormous training time. Additionally, they also exhibit high sensitivity to poor conditioning, meaning that even a tiny change in input can significantly alter the gradient \cite{novak2018sensitivity}. To overcome such a drawback, SGD with momentum (SGDM) \cite{qian1999momentum} has been developed. It introduced first-order momentum to suppress the oscillation of SGD during training and thus made the training more robust. 

It is noted that SGD or SGDM updates parameters using a fixed learning rate. In contrast, the adaptive gradient (AdaGrad) algorithm  \cite{duchi2011adaptive} interpolated second-order momentum which accumulates second-order gradients to achieve an adaptive learning rate. As the number of updates increases, the second-order momentum enables the accumulation of sufficient knowledge,  necessitating a lower learning rate to avoid excessive influence from individual samples. However, the consistently decreasing learning rate may prematurely terminate the training process, preventing the acquisition of essential knowledge from subsequent data. Then the root mean squared propagation (RMSProp) \cite{tieleman2012lecture} has been designed to mitigate the issue by preventing momentum from accumulating all previous gradients. It employs an exponential weighting technique to balance the distant historical information and the knowledge of the current second-order gradients.

As an extensively used tool in DL applications, adaptive moment (Adam) estimation  \cite{kingma2014adam} combines the first-order momentum and adaptive learning rates, thereby delivering robustness to hyperparameters. The update rule for individual weights scales their gradients inversely proportional to the $L_2$ norm of their current and past gradients.  When replacing the $L_2$ norm with an infinite norm, the authors in \cite{kingma2014adam} obtained Adamax which generally exhibits more stable behavior.   In \cite{dozat2016incorporating},  a Nesterov-accelerated adaptive moment (NAdam) estimation was developed to enhance Adam by incorporating the tactic of the Nesterov-accelerated gradient method. For some other gradient-based algorithms, we refer to a survey \cite{ruder2016overview} and the references therein.

\subsection{ADM and ADMM-based learning algorithms} 
Given the capability to decompose a large-scale problem into manageable sub-problems,  alternating direction methods (ADMs) and ADMM \cite{boyd2011distributed} present appealing tools for addressing challenges in distributed manners, making them promising for DL applications such as image compressive sensing \cite{yang2018admm}, federated learning \cite{zhou2023federated}, reinforcement learning  \cite{xu2023federated}, few-shot learning  \cite{wang2023new}, and so forth.

In prior work of \cite{carreira2014distributed, gotmare2018decoupling}, neural network models were divided into sets of layers or blocks that satisfy a consistency constraint, ensuring the output of one set of layers matches the input of the next. The constrained model was relaxed by penalizing these constraints, resulting in an unconstrained penalty model that can be solved by ADMs effectively.

Besides ADMs, a separate line of research explored ADMM to process the neural network models. For instance,   \cite{taylor2016training} relaxed the neural network model by penalizing all equality constraints. Based on the penalty model, a single Lagrange multiplier was added only for the outermost layer, which differs from the standard augmented Lagrange function \cite{boyd2011distributed}. Then ADMM was designed to solve the problem.  \cite{wang2019admm} also penalized all equality constraints except for a linear equation for the outermost layer. A dlADMM algorithm was then proposed to solve the new penalized model. To avoid the computation of matrix inverses,   quadratic approximation was cast to tackle a large-scale system of equations. Similar work can be found in \cite{wang2021convergent}.  Fairly recently,  \cite{ebrahimi2024aa}  employed an Anderson acceleration to boost the convergence rate of dlADMM. Moreover,   \cite{zeng2021admm} leveraged the gradient-free feature of ADMM to develop a sigmoid-ADMM algorithm which enabled mitigating the saturation issue arising from the use of sigmoid activations. This algorithm demonstrated superior performance compared to conventional SGD methods typically used for ReLU-based networks.

We highlight that all these algorithms have been developed based on the neural network model or its reformulations, and thus can be deemed model-driven approaches. Numerical experiments have demonstrated that they enabled faster convergence and better generalization performance across various applications than traditional DL methods. In this work, differing from all prior work, we aim to design a novel data-driven ADMM algorithm.

\subsection{Our contribution}
The primary contribution of this paper lies in the development of an effective data-driven algorithm, BADM, with several advantageous properties.   

\begin{itemize}[leftmargin=10pt]
\item The algorithmic framework is simple but general enough, offering great flexibility to deal with a wide range of applications including various distributed optimization problems and deep learning models such as DNNs, Transformers, multilayer perceptron (MLP), convolutional neural networks (CNNs), graph neural networks (GNNs), U-Net, to name a few. 
\item We prove that the proposed algorithm achieves a sublinear convergence rate in terms of the following form
$$\min_{k=1,2,\ldots, K}  \|\nabla F(\bz^k) \|^2 
= O( \delta+ 1/K),$$
where $F$ is the total loss function, $\nabla F$ is its gradient, $\bz^k$ is the point generated by BADM at the $k$th step, and $\delta$ is related to the sampling error when approximating the gradient. It is worth mentioning that our convergence analysis diverges from those typically employed for SGD-based algorithms.  Moreover, to derive the convergence result outlined in Theorem \ref{lemma:complexity}, we only assume the Lipschitz continuity of the gradient (commonly referred to as L-smoothness in the literature) and the boundedness of sampling error $\delta$. 
\item Distinct from the conventional ADMM paradigms, such as dlADMM \cite{wang2021convergent} and sigmoid-ADMM \cite{zeng2021admm}, which are developed based on the neural network models,  our proposed algorithm can be deemed as a data-driven method. Specifically, we partition the training data into a series of batches and further subdivide each batch into multiple sub-batches.  Based on this data partitioning, we construct an optimization model, as presented in models (\ref{opt-prob}) and (\ref{opt-prob-distribute}), to develop BADM. A detailed comparison of BADM with other ADMM algorithms is provided in Section \ref{com-admm}.

\item  The proposed algorithm enables parallel computing for sub-problems using sub-batch data, resulting in low computational complexity. Extensive numerical experiments on applications in graph modelling, computer vision, image generation, and NLP have demonstrated the high performance of BADM. To be more precise, it achieved higher testing accuracy in most classification tasks, improved training efficiency for image generation models by 3.2 times, and reduced pre-training computation time for language modelling by up to 4 times.
 
\end{itemize}

\subsection{Organization}

The paper is organized as follows. In Section \ref{sec:alg}, we introduce the optimization model based on data partitioning and develop the BADM algorithm, along with its convergence analysis.  Section \ref{com-admm} provides a detailed comparison of BADM with other ADMM algorithms. Section \ref{sec:num} presents comprehensive numerical experiments across four tasks: graph modelling, computer vision, image generation, and  NLP. Concluding remarks are given in the last section.

\section{The BADM Algorithm}\label{sec:alg}
In this section, we begin by introducing the notation that will be used throughout the article and then go through the model formulation and algorithmic design. 

\subsection{Notation}
Throughout the paper, scalars are represented using plain letters, vectors are denoted with bold letters, and matrices are indicated with bold capital letters. We define three sets $\mathbb{B}:=\{1,2,\ldots,B\}$, $\N:=\{1,2,\ldots,N\}$, $\S:=\{1,2,\ldots,S\}$, and denote $\M:=\BB\times\N_b$. Here, `$:=$' means `define'. We 
 use $b$ and $n$ to  indicate their elements, namely $b\in\mathbb{B}, s\in\S$, and $(b,s)\in\M$ . Here $:=$ means define. The cardinality of a set $\D$ is written as $|\D|$. 
For two vectors $\bw$ and $\bpi$, their inner product is denoted  by $\langle\bw,\bpi\rangle:=\sum_iw_i\pi_i$.  Let ${\|\cdot\|}$ be the Euclidean norm, namely, $\|\bw\|^2={\langle\bw,\bw\rangle}$.  
In the sequel, subscripts $i$, $b$, and $s$ respectively represent the index of a sample, a batch, and a sub-batch (e.g., $\bx_i$, $\D_b$, and $\D_{bs}$). We use superscript $\ell$ to stand for the iteration number (e.g., $\bw_{b}^{\ell}$ and $\bw_{bs}^{\ell}$).
\begin{figure*}[!th]
\centering
	\includegraphics[width=\linewidth]{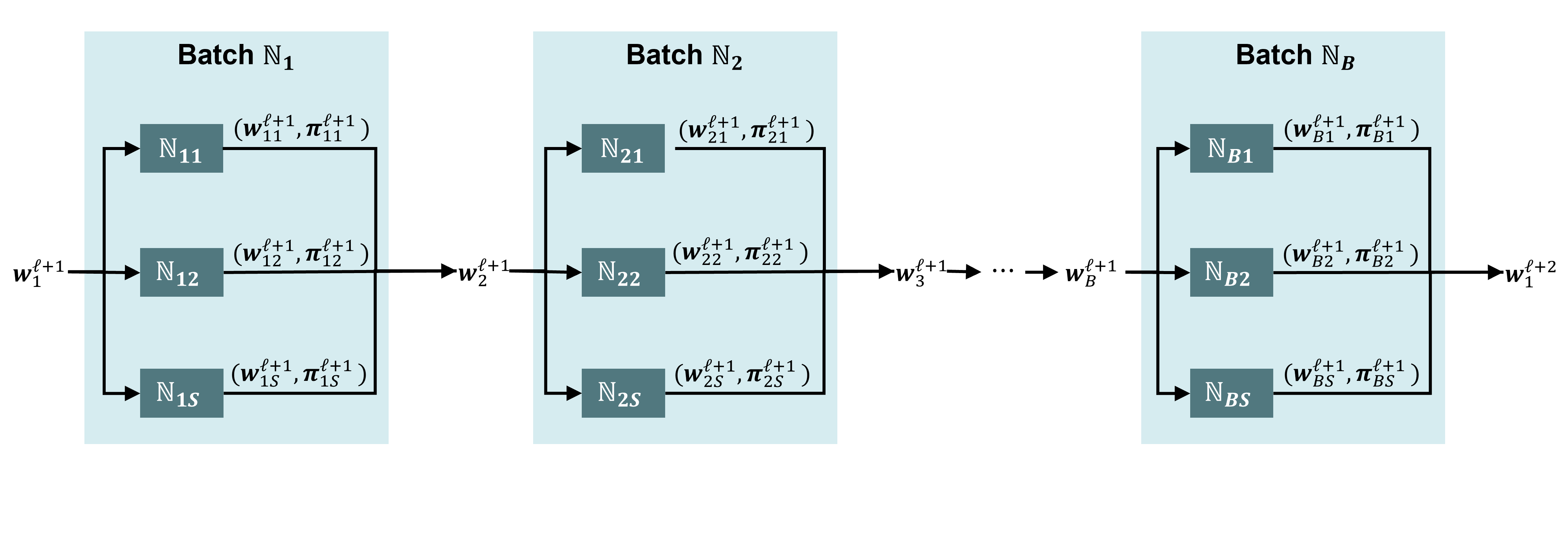}\vspace{-5mm}  
\caption{Data split and each epoch of BADM. \label{fig:train}}
\end{figure*}
\subsection{Model Description}
Suppose we are given a set of date as $\D:=\{(\bx_i,\by_i):i=1,2,\ldots,N\}$, where $\bx_i$ and $\by_i$ are the input and output/label of the $i$-th sample, and $N$ is the total number of samples. Recall that $\N:=\{1,2,\ldots,N\}$ is the indices of all samples. The loss function on this set of data is defined by,
\begin{eqnarray}\label{def-loss-fun}
F(\bw;\D):=\frac{1}{N} \sum_{i\in\N} l\left(f(\bw; \bx_i), \by_i\right),
\end{eqnarray}
where  $l(\cdot)$ is a loss function, $f(\bw;\D)$ is a function (such as linear functions or neural networks) parameterized by $\bw$ and sampled by $\D$. As presented in Figure \ref{fig:train}, we first divide total data indices $\N$ into $B$ disjoint batches, namely,  $\N= \N_1\cup\N_2\cup\ldots\cup\N_B$ and $\N_b\cap\N_{b'}=\emptyset$ for any two distinct $b$ and $b'$. Differing from the standard settings designed for SGD algorithms, in the sequel, we introduce a distributed learning scheme. To proceed with that, as shown in Figure \ref{fig:train}, we further separate batch  $\N_b$ into $S$ disjoint sub-batches, that is, $\N_b= \N_{b1}\cup \N_{b2}\cup\ldots\cup \N_{bS}$  and $ \N_{bs}\cap \N_{bs'}=\emptyset$ for any two distinct $s$ and $s'$. 
By denoting
\begin{eqnarray}  \label{def-Fbn}
\eqspace{1.5}
\begin{array}{rl}
\alpha_{bs}&:= \frac{|\N_{bs}|}{|\N|}=\frac{|\N_{bs}|}{N},\\
\alpha_{s}&:=\sum_{b\in\BB}\alpha_{bs},\\
F_{bs}(\bw)&:=  F_{bs}(\bw;\N_{bs}):=\frac{1 }{|\N_{bs}|}  \sum_{i\in\N_{bs}} l\left(f(\bw; \bx_i), \by_i\right),
\end{array}
\end{eqnarray} 
we can rewrite $F(\bw;\D)$ as follows,
\begin{align*}  
F(\bw;\D) 
 &= \frac{1}{N} \sum_{b\in\BB}  \sum_{s\in\S} \sum_{i\in\N_{bs}} l\left(f(\bw; \bx_i), \by_i\right)\nonumber\\
 &= \sum_{b\in\BB}  \sum_{s\in\S} \alpha_{bs}  F_{bs}(\bw).
\end{align*}
Note that $\sum_{b\in\BB} \sum_{s\in\S} \alpha_{bs}=1$.
Overall, the goal is to learn an optimal parameter $\bw^*$ by solving
\begin{eqnarray}  \label{opt-prob}
\bw^*= {\rm arg}\min\limits_{\bw} F(\bw;\D)   = {\rm arg}\min\limits_{\bw}  \sum_{b\in\BB}  \sum_{s\in\S} \alpha_{bs}  F_{bs}(\bw). 
\end{eqnarray}
In the paper, we always assume that the optimal function value is  bounded from below, that is, $$F^*:=F(\bw^*;\D) >-\infty.$$

\subsection{The algorithmic design}

In order to adopt the ADMM algorithm, we rewrite problem (\ref{opt-prob}) as follows,
\begin{eqnarray}  \label{opt-prob-distribute}
\begin{aligned}
\min\limits_{\bw, \bw_{bs}}&~~\sum_{b\in\BB} \sum_{s\in\S} \alpha_{bs}  F_{bs}(\bw_{bs}),\\
{\rm s.t. }&~~\bw = \bw_{bs},~ b\in\BB,~s\in\S.
\end{aligned}
\end{eqnarray}
Besides the global parameter $\bw$, additional variables $\bw_{bs}$ (i.e., the local parameter for sub-batch $\N_{bs}$) are introduced in the above model. The corresponding augmented Lagrange function  is,
\begin{eqnarray*}  \label{opt-prob-distribute-Lag}
\begin{aligned}
\L\left(\bw,\{(\bw_{bs},\bpi_{bs}):(b,s)\in\M\}\right)\\:=\sum_{b\in\BB}  \sum_{s\in\S}  \alpha_{bs} \L_{bs}(\bw,\bpi_{bs}, \bw_{bs}),
\end{aligned}
\end{eqnarray*}
where  $\L_{bs}$ is given by
\begin{eqnarray*} 
\begin{aligned}
&\L_{bs}(\bw, \bw_{bs},\bpi_{bs})\\
:=  & F_{bs}(\bw_{bs}) + \langle \bpi_{bs}, \bw_{bs}- \bw\rangle  + \frac{\sigma}{2} \|\bw_{bs}- \bw\|^2.
\end{aligned}
\end{eqnarray*}
and ${\sigma>0}$. Here ${\{\bpi_{bs}:(b,s)\in\M\}}$ are the Lagrange multipliers.  
The framework of ADMM is described as follows. By initializing ${(\bw^0,\{(\bw_{bs}^0,\bpi_{bs}^0):(b,s)\in\M\})}$, we perform the following steps iteratively for each epoch $\ell\in\{0,1,2,\ldots\}$:
For $b=0$,
$$\bw_{0s}^{\ell+1}=\bw_{Bs}^{\ell},~~\bpi_{0s}^{\ell+1}=\bpi_{Bs}^{\ell},\qquad s\in\S.$$
For each $b\in\BB$,
\begin{subequations}
\label{frame-ADMM}	
			\begin{alignat}{4}	
			 \label{wb-update}	
			\bw_b^{\ell+1} &= {\rm arg}\min_{\bw} \sum_{s\in\S}  \alpha_{bs} \L_{bs}(\bw, \bw_{(b-1)s}^{\ell+1},\bpi_{(b-1)s}^{\ell+1}),\\[1ex]			\label{wbn-update}
			\bw_{bs}^{\ell+1} &= {\rm arg}\min_{\bw_{bs}} \L_{bs}(\bw_{b}^{\ell+1}, \bw_{bs},\bpi_{(b-1)s}^{\ell+1}), ~~s\in\S,\\[1.25ex]
			 \label{pibn-update}	
			\bpi_{bs}^{\ell+1} &= \bpi_{(b-1)s}^{\ell+1} + \sigma(\bw_{bs}^{\ell+1}- \bw_{b}^{\ell+1}),\hspace{9.5mm}s\in\S.
			 \end{alignat}
		\end{subequations}
For sub-problems of $\bw_b$ in \eqref{wb-update}, it is solved by
\begin{align*}
\bw_{b}^{\ell+1}&= {\rm arg}\min\limits_{\bw} \sum_{s\in\S}\alpha_{bs}\Big(\frac{\sigma}{2} \|\bw_{(b-1)s}^{\ell+1}- \bw\|^2-\langle \bpi_{(b-1)s}^{\ell+1},  \bw\rangle \Big).
\end{align*}
However, instead of solving the above problem directly, we aim to address the following problem, 
\begin{align}\label{sub-wb-secod}
\bw_{b}^{\ell+1}&= {\rm arg}\min\limits_{\bw_{b}} \sum_{s\in\S}\alpha_{s}\left(\frac{\sigma}{2} \|\bw_{(b-1)s}^{\ell+1}- \bw_{b}\|^2-\left\langle {\bpi}_{(b-1)s}^{\ell+1},  \bw_{b}\right\rangle \right)\nonumber\\
&=\sum_{s\in\S}\alpha_{s}\left(\bw_{(b-1)s}^{\ell+1}+   \frac{ {\bpi}_{(b-1)s}^{\ell+1}}{\sigma}\right).
\end{align}
The only difference between the above two problems lies in using the weights, i.e., $\alpha_{bs}$ and $\alpha_{s}$. Our numerical experiments have demonstrated the similar performance of using $\alpha_{bs}$ and $\alpha_{s}$. However, exploiting the latter leads to the ease of convergence analysis.
To accelerate the computational speed, we solve  sub-problems of $\bw_{bs}$ in \eqref{wbn-update} inexactly by
\begin{eqnarray}\label{sub-wbn}
\begin{aligned}
\bw_{bs}^{\ell+1} &= {\rm arg}\min\limits_{\bw_{bs}} ~ \langle    \bpi_{(b-1)s}^{\ell+1}, \bw_{bs} \rangle + \frac{ \sigma}{2}\|\bw_{bs} -\bw_{b}^{\ell+1}\|^2 \\ 
&+ \langle  \nabla F_{bs}(\bw_{b}^{\ell+1})  , \bw_{bs} \rangle + \frac{\rho}{2}\|\bw_{bs} -\bw_{b}^{\ell+1} \|^2 \\ 
&=\bw_{b}^{\ell+1} - \frac{  \nabla F_{bs}(\bw_{b}^{\ell+1})+\bpi_{(b-1)s}^{\ell+1}}{\rho +\sigma},
\end{aligned}
\end{eqnarray}
where ${\rho>0}$ and $\nabla F_{bs}(\bw)$ is one of elements in the sub-differential (denoted by $\partial F_{bs}(\bw)$, see \cite[Definition 8.3]{rockafellar2009variational}, of $F_{bs}(\bw)$. Note that $\nabla F_{bs}(\bw)$ is the gradient of $F_{bs}(\bw)$ if it is continuously differentiable at $\bw$. 

 The overall algorithmic framework is presented in Algorithm \ref{algorithm-ADMM-sketch}. The process of each epoch is illustrated in Figure \ref{fig:train}. Its advantageous properties are highlighted as follows.

\begin{algorithm}[!t]
    \SetAlgoLined    
{\justifying\noindent Divide $\D$ into $B$ disjoint  batches $\{\N_1,\ldots,\N_B\}$ and split  $\N_b,~b\in\BB$ into $S$ disjoint sub-batches $\{\N_{b1}, \ldots, \N_{bS}\}$. Calculate  $\alpha_{bs}$ and $\alpha_{s}$  by (\ref{def-Fbn}) for  $(b,s)\in\M$. Initialize $(\sigma, \rho, L)>0$, $\beta\in(0,1)$, and $\{(\bw_{bs}^0,\bpi_{bs}^0):(b,s)\in\M\}$. }

\For{$\ell=0,1,2,\ldots,L$}{
Set ${(\bw_{0s}^{\ell+1},\bpi_{0s}^{\ell+1})=(\bw_{Bs}^{\ell},\bpi_{Bs}^{\ell}})$ for each $s\in\S$.

\For{$b=1,2,\ldots,B$}{ 

 Update $\bw_{b}^{\ell+1}$ by (\ref{sub-wb-secod}).
 
\For{$s=1,2,\ldots,S$}{
Update $(\bw_{bs}^{\ell+1}, \bpi_{bs}^{\ell+1}) $ by  (\ref{sub-wbn}) and (\ref{pibn-update}).
}  
}
} 
Return $\bw_B^{\ell+1}$.
\caption{Batch ADMM (BADM) }\label{algorithm-ADMM-sketch}
\end{algorithm}

 \begin{itemize}[leftmargin=10pt]
  \item \textbf{Comparable computational complexity.} The primary computational expense in Algorithm \ref{algorithm-ADMM-sketch} arises from computing $\nabla F_{bs}(\bw^{\ell})$ for each sub-batch $\N_{bs}$. Consequently, its computational burden resembles that of standard SGD-based algorithms. Hence, Algorithm \ref{algorithm-ADMM-sketch} does not exhibit higher computational complexity compared to most widely-used algorithms.

   \item  \textbf{Parallel computing.} Within each batch $b$, $S$ blocks of parameters $(\bw_{b1}^{\ell}, \bpi_{b1}^{\ell}),\ldots,(\bw_{bS}^{\ell}, \bpi_{bS}^{\ell})$ can be computed in parallel, enabling fast computation.

 \end{itemize}
 
\subsection{Convergence analysis}\label{sec:convergence}
To establish the convergence results for BADM in Algorithm \ref{algorithm-ADMM-sketch}, we need the following assumption.
\begin{assumption}\label{assum-Lip} For a given $(\bx, \by)$, the gradient of $l\left(f(\cdot; \bx), \by\right)$ is Lipschitz continuous with a constant $\eta(\bx, \by)>0$, namely,
$$\|\nabla l\left(f(\bw; \bx), \by\right)-\nabla l\left(f(\bv; \bx), \by\right)\|\leq \eta(\bx, \by) \|\bw-\bv\|.$$
\end{assumption}
The above condition is known as the L-smoothness, which is commonly assumed to establish the convergence property for non-convex optimization problems. Hereafter, we define two useful constants by
\begin{equation}\label{def-eta-delta}
\begin{aligned}
\eta&:=\sup_{(\bx_i, \by_i)\in\D}\eta(\bx_i, \by_i),\\
\delta&:=\sup_{s\in\S}\sup_{b\in\BB}\sup_{\bw}100\|\nabla F_{s}(\bw;\D_s)  -\nabla F_{bs}(\bw;\N_{bs})\|^2,
\end{aligned}
\end{equation} 
and always set
\begin{align}
\label{lower-bd-sigma}
\sigma\geq\max\{5\eta,5\rho\}, 
\end{align}
where $F_{bs}$ is defined by \eqref{def-Fbn} and $F_{s}$ is defined by 
\begin{align}
\label{def-Fs}
F_{s}(\bw)&:=F_{s}(\bw;\D_s):=\frac{\sum_{b\in\BB} \sum_{i\in\N_{bs}} l\left(f(\bw; \bx_i), \by_i\right)}{\sum_{b\in\BB}|\N_{bs}|},\\[1ex]
 \label{def-Ds}
\D_s&:= \N_{1s}\cup\N_{2s}\ldots\cup\N_{Bs},~~ s\in\S.
\end{align}
Based on the definition of $F_{s}$, one can see that
\begin{align} \label{F-Fs-alpha}
F(\bw;\D) =\sum_{s\in\S }\alpha_s F_{s}(\bw;\D_s)= \sum_{s\in\S }\alpha_s F_{s}(\bw).
\end{align}
The second assumption given below is the boundedness of $\delta$. This constant is related to the sampling bias, which is frequently assumed to be bounded in \cite{stich2018local, Li2020On, CooperativeSGD21, Adan24}. 
\begin{assumption}\label{assum-bias} Suppose that $\delta <\infty.$
\end{assumption} 
\noindent We note that if $B=1$ then $\D_s= \N_{1s}$ for each $s\in\S$, thereby $\delta=0$. Under such a case, the above assumption holds automatically. For notational convenience,  we denote
\begin{eqnarray}\label{def-z-Z-pi}
\begin{aligned}
k&:=\ell B+b,\\
\bz^k&:=\bz^{ \ell B+b} :=  \bw^{\ell+1}_b, &b\in\BB,~&\\
\bz_s^k&:=\bz_s^{\ell B+b} :=  \bw^{\ell+1}_{bs},&b\in\BB,~&s\in\S,\\
\bv_s^k&:=\bv_s^{\ell B+b} :=  \bpi^{\ell+1}_{bs},&b\in\BB,~&s\in\S.
\end{aligned}
\end{eqnarray}
Our first result shows a descent property. If $B=1$ then $\delta=0$, leading to a strict descent property. 
\begin{lemma}\label{lemma:descent} Let ${(\bw^\ell,\{(\bw_{bs}^\ell,\bpi_{bs}^\ell):(b,s)\in\M\})}$ be the sequence generated by BADM. Under Assumptions \ref{assum-Lip} and \ref{assum-bias}, 
\begin{equation} \label{decrease-L}
\begin{aligned}
\L^k -\L^{k-1} \leq \frac{\delta}{2\sigma }- \frac{ \sigma}{10} \sum_{s\in\S} \alpha_s\left( \|\bz_{s}^{k}-\bz^{k}\|^2+\|\bz^{k} -\bz^{k-1}\|^2\right),
\end{aligned}
\end{equation}
where $\L^k$ is defined by
\begin{equation}  \label{def-Lk-lemma}
\begin{aligned}
\L^k&:= \sum_{s\in\S}\alpha_s\left(F_{s}(\bz_{s}^k) + \langle \bv_{s}^k, \bz_{s}^k- \bz^k\rangle  + \frac{\sigma}{2} \|\bz_{s}^k- \bz^k\|^2\right).
\end{aligned}
\end{equation} 
\end{lemma}
\begin{theorem}\label{lemma:complexity} Let ${(\bw^\ell,\{(\bw_{bs}^\ell,\bpi_{bs}^\ell):(b,s)\in\M\})}$ be the sequence generated by BADM. Under Assumptions \ref{assum-Lip} and \ref{assum-bias}, 
\begin{equation*}  
\begin{aligned}
\min_{k=1,2,\ldots, K}  \|\nabla F(\bz^k) \|^2 
\leq 12\delta +  \frac{ 24\sigma(\L^0- F^*)}{K}.
\end{aligned}
\end{equation*}
\end{theorem}
In the above theorem, error $\delta $ stems from the sampling when calculating gradient $\nabla F_{s}(\cdot;\D_s)$ which is approximated by using $\nabla F_{bs}(\cdot;\N_{bs})$.  Consequently, this error $\delta$ 
 is unavoidable. In particular, when ${B=1}$,  it follows {$\delta=0$}. In this case,   the iteration complexity reduces to $O(1/K)$, a sublinear rate. 
\section{Comparison with other algorithms}\label{com-admm}
The conventional neural network  model takes the form of 
\begin{eqnarray}\label{DNN-model}
\begin{aligned}
\min_{\W,\V}&~~  l(\V_{N}, \Y),\\
{\rm s.t.} &~~ \V_{i}=\varphi_i(\W_i\V_{i-1}), i=1,2,\ldots,N,
\end{aligned}
\end{eqnarray}
where  $l$ is the loss function, e.g., the average mean squared error \cite{zeng2021admm}, $\varphi_i$ is the activation function for the $i$-th layer (e.g., ReLU or sigmoid for inner layers and softmax or linear functions for the outermost layer), $\W:=(\W_1,\W_2,\ldots,\W_{N})$, $\V=(\V_0,\V_1,\ldots,\V_{N})$, $ \Y:=(\by_1,\by_2,\ldots,\by_{N})$, and $ \V_0:=(\bx_1,\bx_2,\ldots,\bx_{N})$. Let $\|\cdot\|_F$ be the Frobenius norm. Then the augmented Lagrange function of the above problem is 
\begin{eqnarray*}  
\eqspace{1.5}
\begin{array}{lll}
\L(\W,\V,\LA) &=& l(\V_{N}, \Y)  + \sum_{i=1}^{N} (\langle \V_{i}-\varphi_i(\W_i\V_{i-1}), \LA_i\rangle \\ 
&+&  \frac{\sigma}{2}\|\V_{i}-\varphi_i(\W_i\V_{i-1})\|_F^2).
\end{array}
\end{eqnarray*}
The scheme of ADMM usually updates $\W_i$ in the backward order as $\W_{N}\to\ldots\to\W_{2}\to\W_{1}$, then updates $\V_i$  in the forward
order as $\V_{1}\to\V_{2}\ldots\to\V_{N}$, and
finally updates multipliers $\LA_i$ in a parallel way. To be more specific, at iteration $\ell$,   $\W_i$,  $\V_i$, and $\LA_i$ are updated by
\begin{eqnarray*} 
\begin{aligned}
\W_i^{\ell+1} &=\argmin_{\W_i} \langle \V_{i}^{\ell}-\varphi_i(\W_i\V_{i-1}^{\ell}), \LA_i^{\ell}\rangle\\ &+  \frac{\sigma}{2}\|\V_{i}^{\ell}-\varphi_i(\W_i\V_{i-1}^{\ell})\|_F^2,~i=N,N-1,\ldots,1,\\
\V_i^{\ell+1} &=\argmin_{\V_i} ~\langle \V_{i}-\varphi_i(\W_i^{\ell+1}\V_{i-1}^{\ell+1}), \LA_i^{\ell}\rangle\\ &+  \frac{\sigma}{2}\|\V_{i}-\varphi_i(\W_i^{\ell+1}\V_{i-1}^{\ell+1})\|_F^2,~i=1,2,\ldots,N,\\
\V_{N}^{\ell+1} &=\argmin_{\V_{N}} ~l( \V_N, \Y) + \langle \V_{N}-\varphi_i(\W_{N}^{\ell+1}\V_{N-1}^{\ell+1}), \LA_{N}^{\ell}\rangle\\ &+ \frac{\sigma}{2}\|\V_{N}-\varphi_i(\W_{N}^{\ell+1}\V_{N-1}^{\ell+1})\|_F^2,\\ 
\LA_i^{\ell+1} &= \LA_i^{\ell} + \sigma(\V_{i}^{\ell+1}-\varphi_i(\W_i^{\ell+1}\V_{i-1}^{\ell+1})),~i=1,2,\ldots,N.
\end{aligned}
\end{eqnarray*}
where  ${\V_0^\ell=\V_0}$ for all $\ell$. We note that both dlADMM \cite{wang2021convergent} and sigmoid-ADMM \cite{zeng2021admm} follow similar structures of the above algorithm. However, our proposed algorithm is fundamentally different from these algorithms. Firstly, the above ADMM framework is based directly on DNN model (\ref{DNN-model}) or its variants, such as the penalty model in \cite{wang2021convergent}. In contrast, BADM is developed based on model (\ref{opt-prob-distribute}), which is driven by data partitioning. Furthermore, in BADM, all $\bw_{bs}$, $\bw_b$, and $\bw$ can be deemed as $\W$ in model (\ref{DNN-model}). Consequently, dlADMM and sigmoid-ADMM update each portion $\W_i$ of $\W$ sequentially in each iteration, whereas BADM treats $\bw$ as a whole entity and updates it all at once in each iteration.

\section{Evaluation of BADM}
\label{sec:num}
This section evaluates the performance of BADM by comparing it with several leading optimizers across four different tasks: graph modelling, computer vision, image generation, and NLP. Specifically, for graph modelling, we compare BADM against six benchmarks: Adam, RMSProp, AdaGrad, SGD, NAdam, and dlADMM. For the other three tasks, we compare BADM with Adam and RMSProp. All optimizers use their default hyperparameters as provided by TensorFlow's built-in functions, without any regularization or decay functions applied during training.

\subsection{Graph modelling}

 We undertake two tasks in graph modelling. The first task is node-level classification, which predicts node categories based on node features and their relationships with other nodes. To evaluate this task, we take advantage of two deep learning structures: DNN and GNN. The second task is graph-level prediction, where we use a message passing neural network (MPNN) to predict the molecular property known as blood-brain barrier permeability (BBBP).

\textbf{a) Node classification.} Following the methodology outlined in \cite{you2020design}, we use identical network structures for both DNN and GNN and only modify the training process. Our evaluation encompasses six benchmark datasets: `Cora', `Pubmed', `Citeseer' \cite{sen2008collective}, `Coauthor CS',  `Coauthor Physics' \cite{shchur2018pitfalls}, and `AMZ Computers'. The first five datasets are extracted from citation networks, where nodes represent publications and edges denote citations. The last dataset, `AMZ Computers', comprises a co-purchase graph from Amazon, where nodes represent products, edges indicate co-purchase relations, and features are bag-of-words vectors from product reviews. For DNN experiments, we use only node features for classification, whereas in GNN experiments, we include relations between publications and products as edge features. The data statistics are similar for all datasets. For example,  `Cora' contains 2,708 scientific papers, each classified into one of seven categories. The citation network includes 5,429 links between these papers. Each paper is represented by a binary word vector of length 1,433, indicating the presence of specific words, resulting in 1,433-dimensional node features. The edge features indicate whether two papers cite each other. 

\begin{table}[t]
\renewcommand{\arraystretch}{1.5}\addtolength{\tabcolsep}{-2.6pt}
    \centering
    \caption{Hyperparameters and testing accuracy for node classification.}
    \label{table:1}
    \begin{tabular}{cccccccc}
        \hline
          & Cora & Citeseer & PubMed & Physics & CS & Computers\\
        \hline
      $B$ & 128 & 512 & 512 & 512 & 512 & 512 \\ 
       $S$ & 16 & 64 & 128 & 128 & 256 & 128 \\ 
       $\rho$ & 200 & 600 & 300& 500& 950 & 400\\ 
       $\sigma$ &800 & 400 & 700 & 500 & 50 & 600\\ 
       \hline
       \multicolumn{7}{c}{DNN}\\\hline
      SGD & 0.3034 & 0.2263 & 0.3852 & 0.6937 & 0.2645 & 0.3708 \\ 
       AdaGrad & 0.7215 & 0.7243 & 0.8411 & 0.9597 & 0.8593 & 0.6859 \\ 
       dlADMM & 0.3428 & 0.4084 & 0.5897 & 0.8138 & 0.5716 & 0.5538 \\ 
       RMSProp & 0.7200 & 0.7334 & 0.8712 & 0.9683 & 0.9176 & 0.8316 \\ 
       Adam & 0.7358 & 0.7263 & 0.8731 & 0.9689 & 0.9094 & 0.8243 \\ 
       NAdam & 0.7343 & 0.7323 & 0.8727 & 0.9688 & 0.9176 & 0.8216 \\ 
       BADM & \textbf{0.7464} & \textbf{0.7424} & \textbf{0.8736} & \textbf{0.9695} & \textbf{0.9185} & \textbf{0.8403} \\ \hline

       \multicolumn{7}{c}{GNN}\\\hline
       SGD & 0.6302 & 0.1982 & 0.3676 & 0.6937 & 0.3007 & 0.3711 \\ 
       AdaGrad & 0.7268 & 0.6771 & 0.8411 & 0.9610 & 0.8523 & 0.6859 \\ 
       RMSProp & 0.7781 & 0.6881 & 0.8871 & \textbf{0.9699} & 0.9412 & \textbf{0.8492} \\ 
       Adam & 0.7842 & 0.6962 & 0.8873 & 0.9686 & 0.9438 & 0.8463 \\ 
       NAdam & 0.7857 & 0.6911 & 0.8864 & 0.9687 & 0.9438 & 0.8476 \\ 
       BADM & \textbf{0.7925} & \textbf{0.7213} & \textbf{0.8922} & {0.9688} & \textbf{0.9495} & 0.8400 \\ \hline

    \end{tabular}%
    
    \medbreak
    \end{table}

 We employ a DNN with two hidden layers and 32 neurons in each layer for these experiments. The GNN model follows the structure in \cite{you2020design}. It begins by preprocessing node features using a DNN (with the same structure as in the previous experiment) to create initial representations. Then, two graph convolutional layers with skip connections are applied to generate node embeddings. Post-processing with a DNN refines these embeddings, which are then passed through a Softmax layer to predict node classes. The learning rate is set to 0.001 for all optimizers, while for BADM we maintain $(\rho,\sigma)$ to satisfy ${ 1/(\rho+\sigma)=0.001}$, such as ${(\rho,\sigma)=(200,800)}$ for dataset `Cora'. Specific hyperparameters for the six datasets are given in Table \ref{table:1}. 
    
    \begin{figure*}[!th]
\centering
\begin{subfigure}{.325\textwidth}
	\centering
	\includegraphics[width=.98\linewidth]{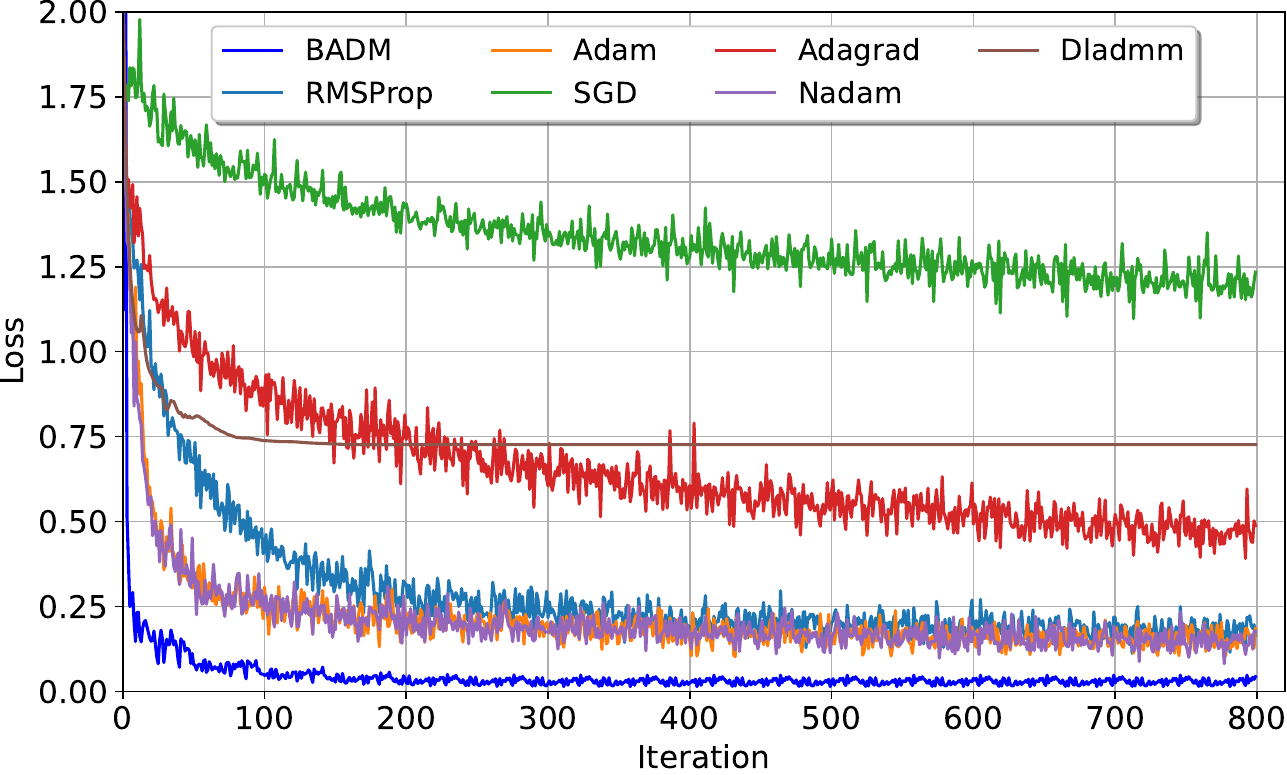} 
	\caption{DNN for Physics}
\end{subfigure}	  
\begin{subfigure}{.325\textwidth}
	\centering
	\includegraphics[width=.98\linewidth]{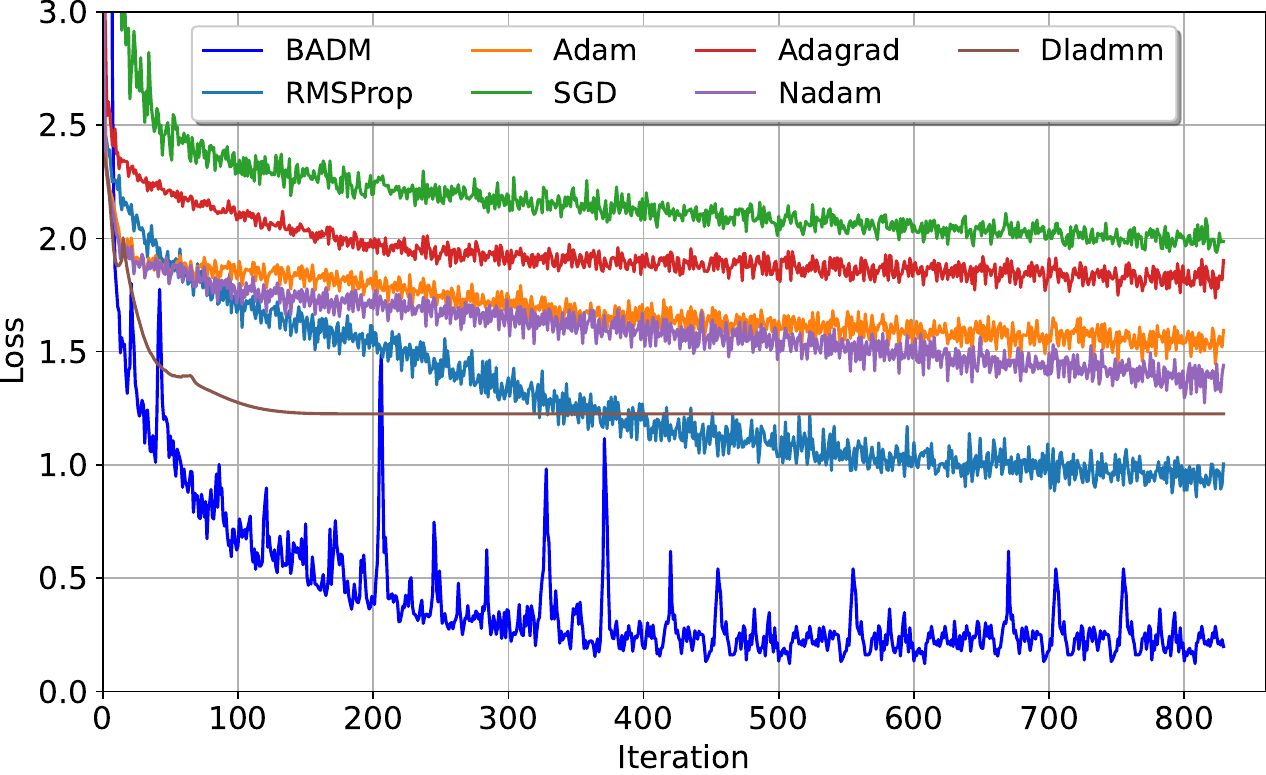} 
	\caption{DNN for Computers}
\end{subfigure}	 
\begin{subfigure}{.325\textwidth}
	\centering
	\includegraphics[width=.98\linewidth]{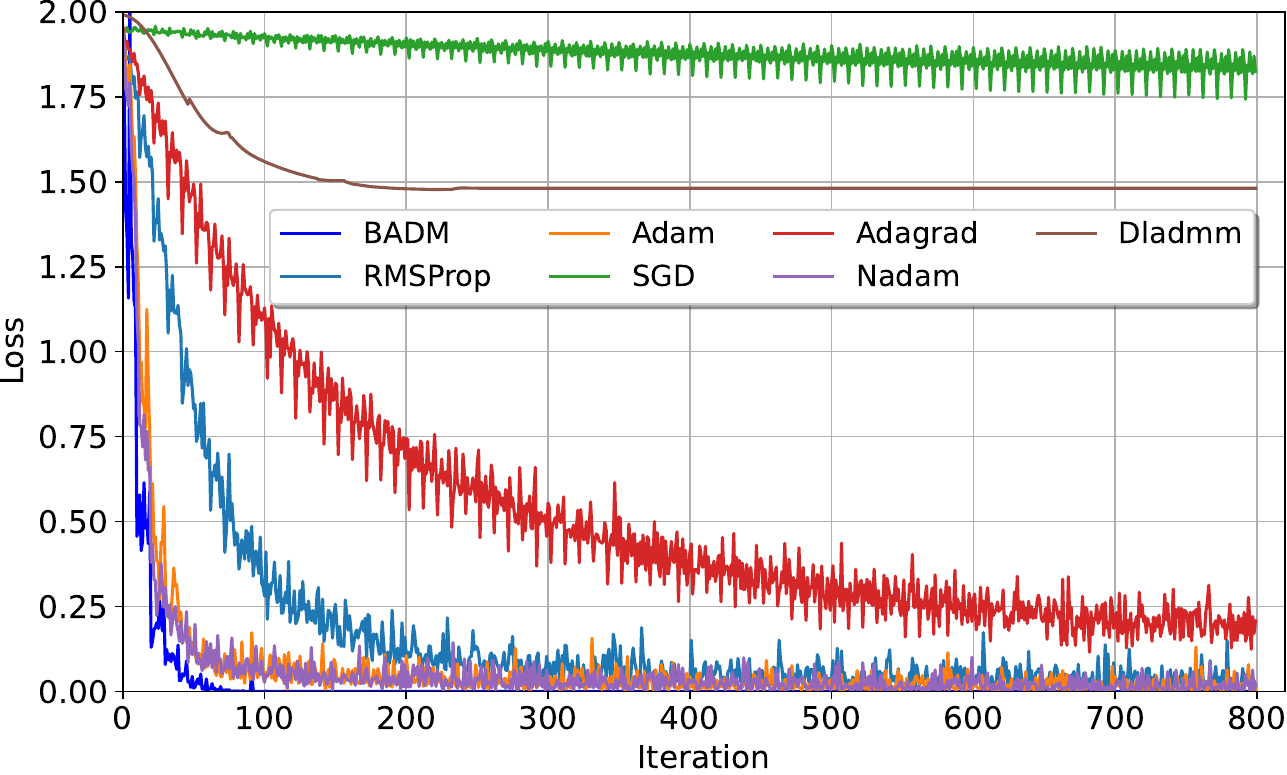} 
	\caption{DNN for Cora}
\end{subfigure}  \\  \vspace{3mm}
\begin{subfigure}{.325\textwidth}
	\centering
	\includegraphics[width=.98\linewidth]{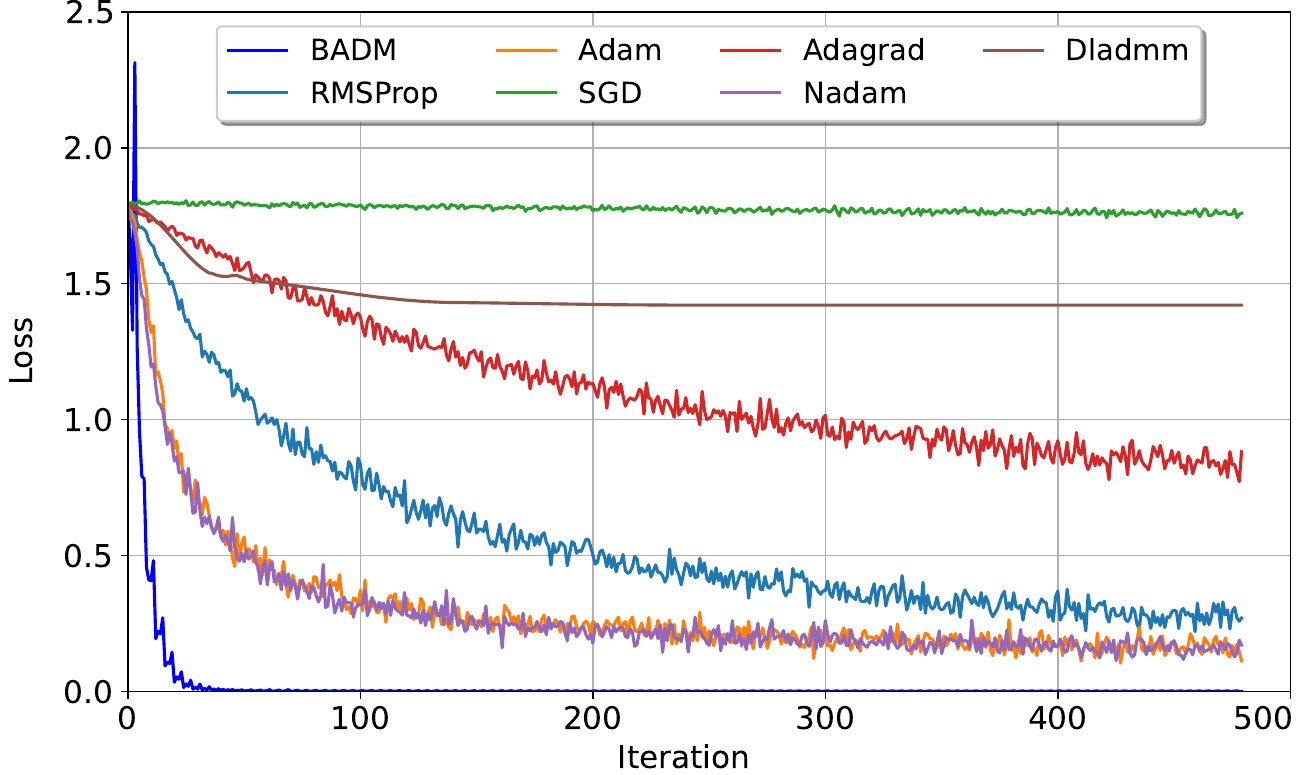} 
	\caption{DNN for Citeseer}
\end{subfigure}
\begin{subfigure}{.325\textwidth}
	\centering
	\includegraphics[width=.98\linewidth]{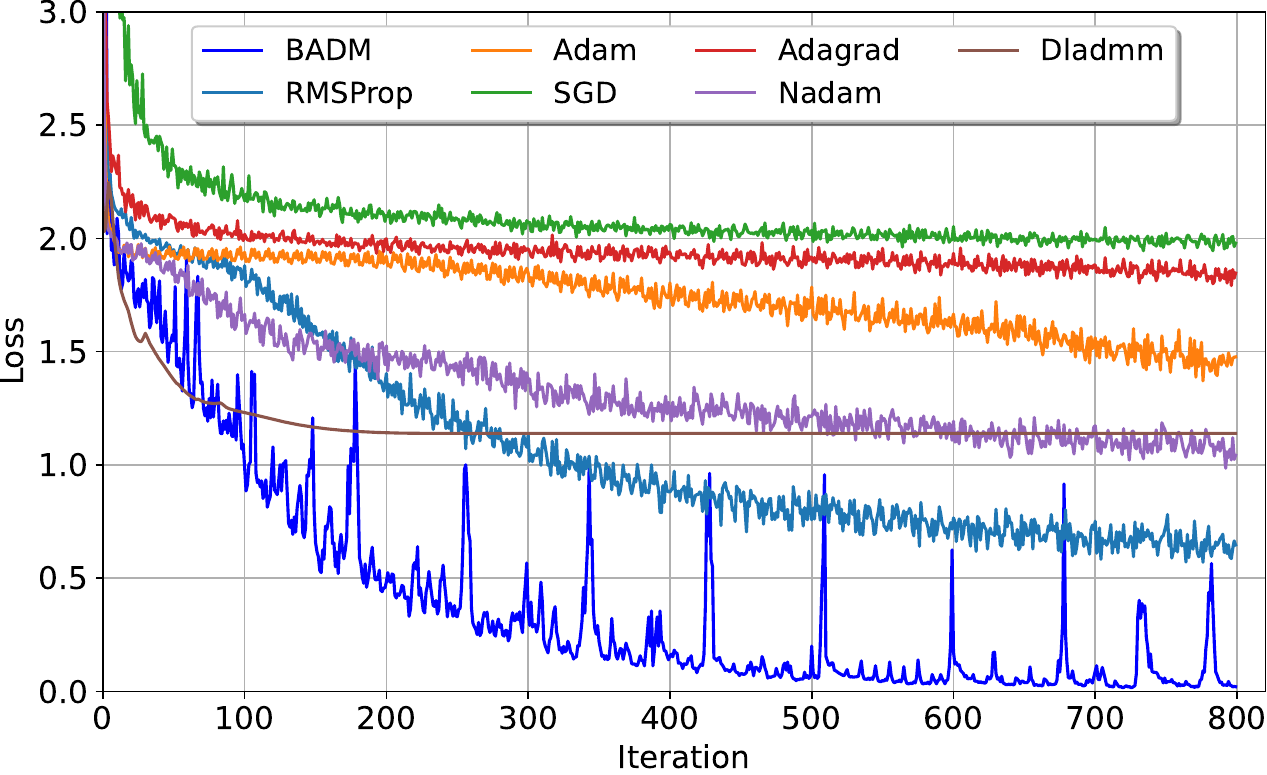} 
	\caption{DNN for CS}
\end{subfigure} 
\begin{subfigure}{.325\textwidth}
	\centering
	\includegraphics[width=.98\linewidth]{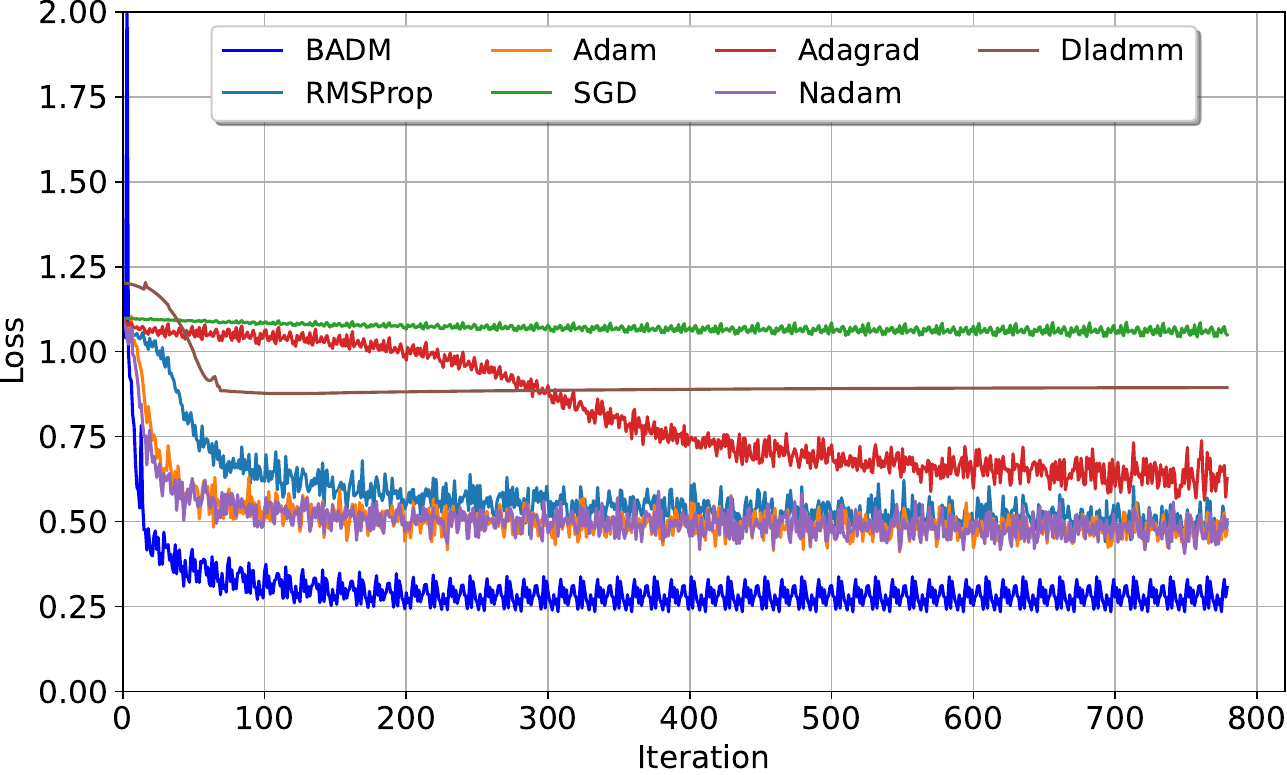} 
	\caption{DNN for Pubmed}
\end{subfigure}\\  \vspace{3mm}

\begin{subfigure}{.325\textwidth}
	\centering
	\includegraphics[width=.98\linewidth]{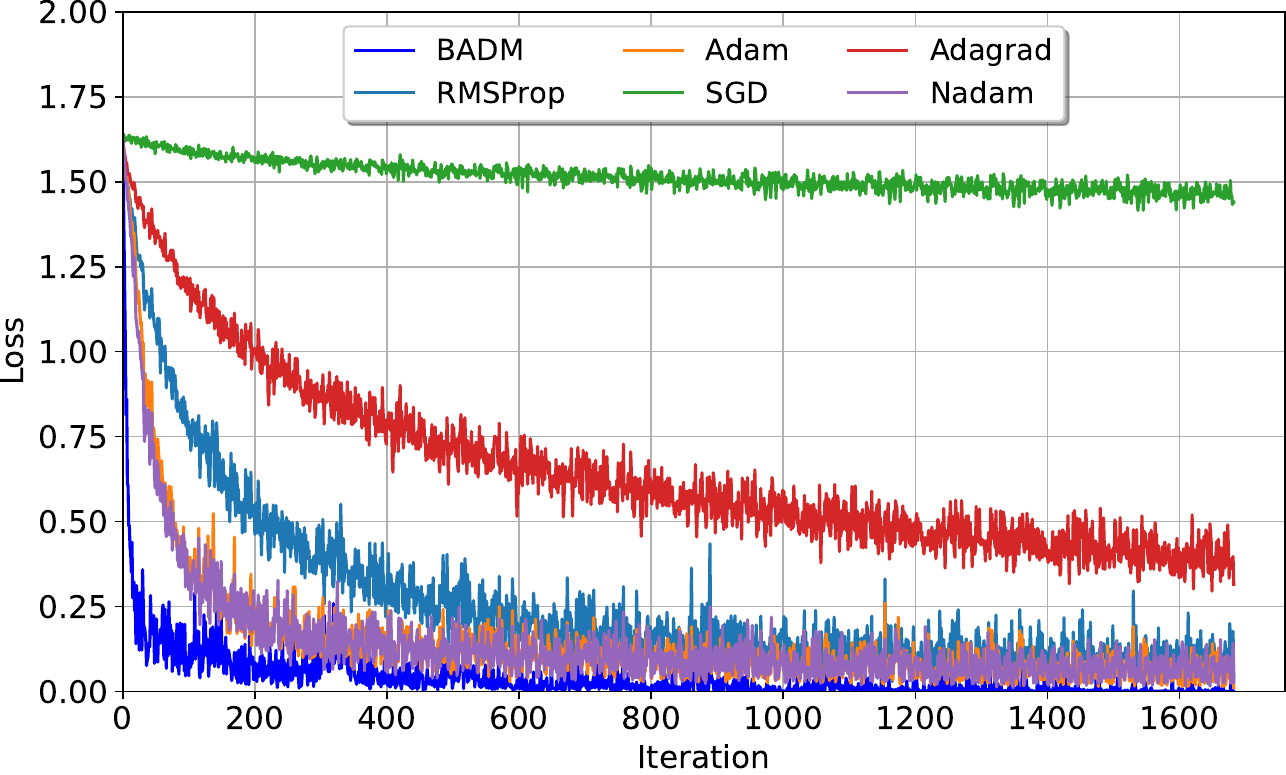} 
	\caption{GNN for Physics}
\end{subfigure}	  
\begin{subfigure}{.325\textwidth}
	\centering
	\includegraphics[width=.98\linewidth]{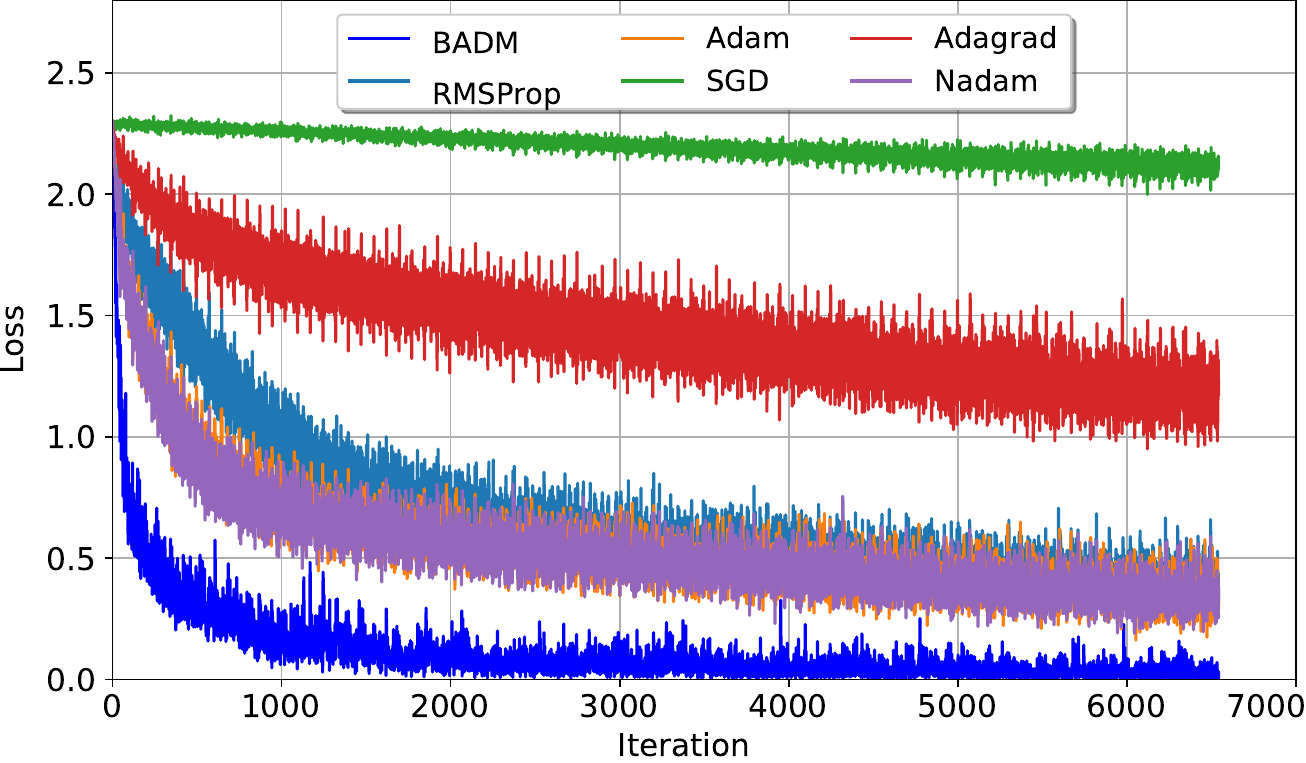} 
	\caption{GNN for Computers}
\end{subfigure}	 
\begin{subfigure}{.325\textwidth}
	\centering
	\includegraphics[width=.98\linewidth]{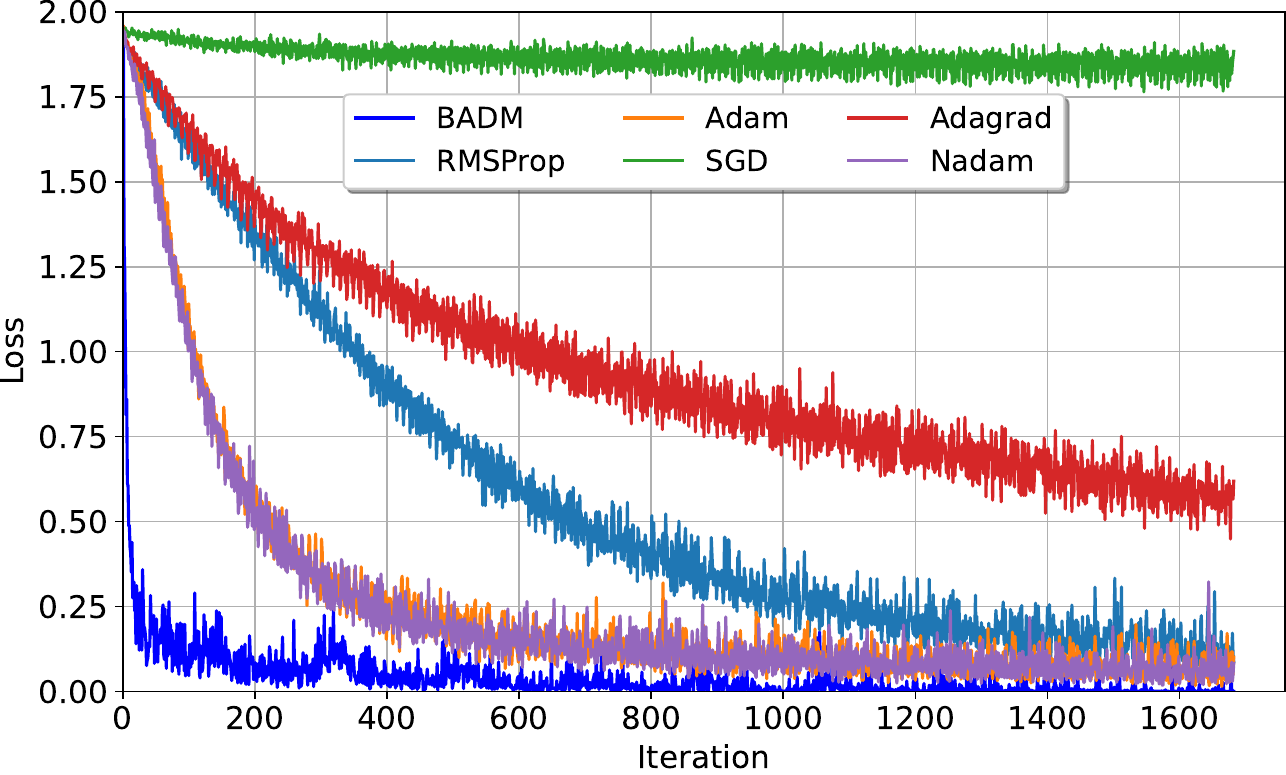} 
	\caption{GNN for Cora}
\end{subfigure} \\  \vspace{3mm}
\begin{subfigure}{.325\textwidth}
	\centering
	\includegraphics[width=.98\linewidth]{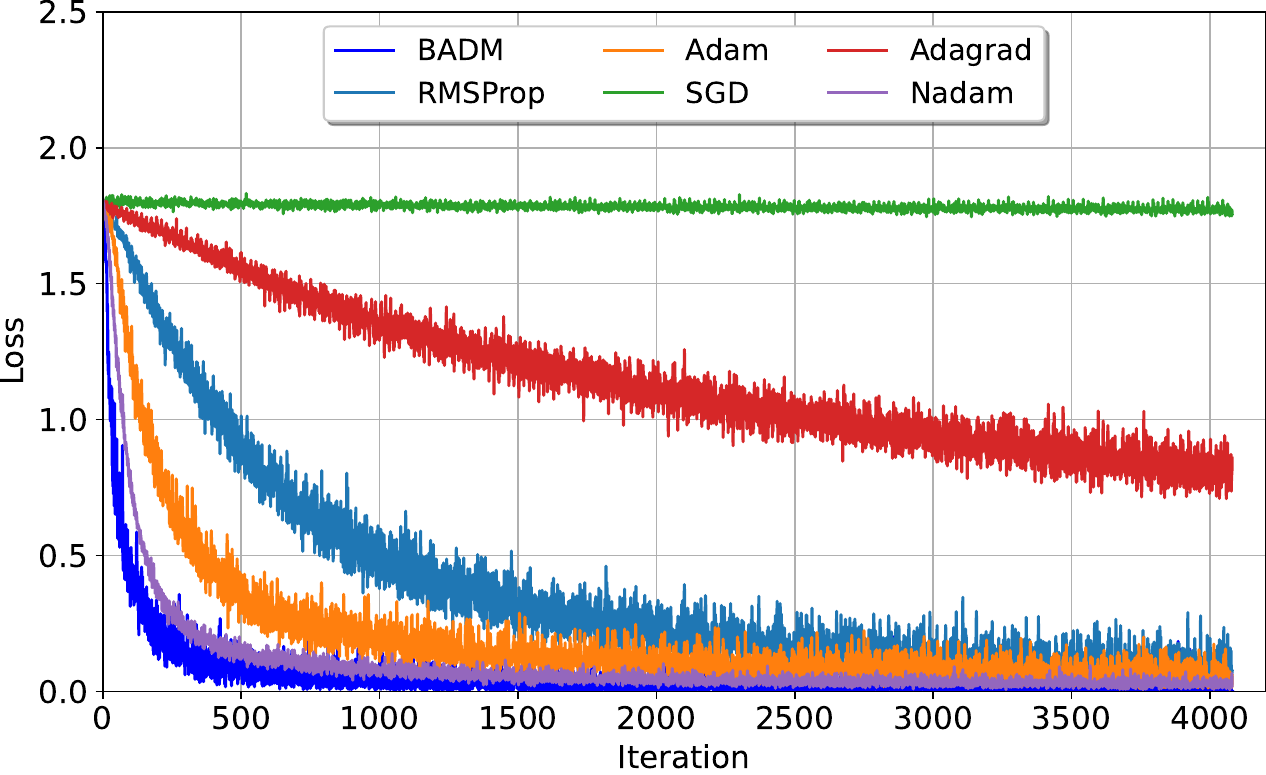} 
	\caption{GNN for Citeseer}
\end{subfigure} 
\begin{subfigure}{.325\textwidth}
	\centering
	\includegraphics[width=.98\linewidth]{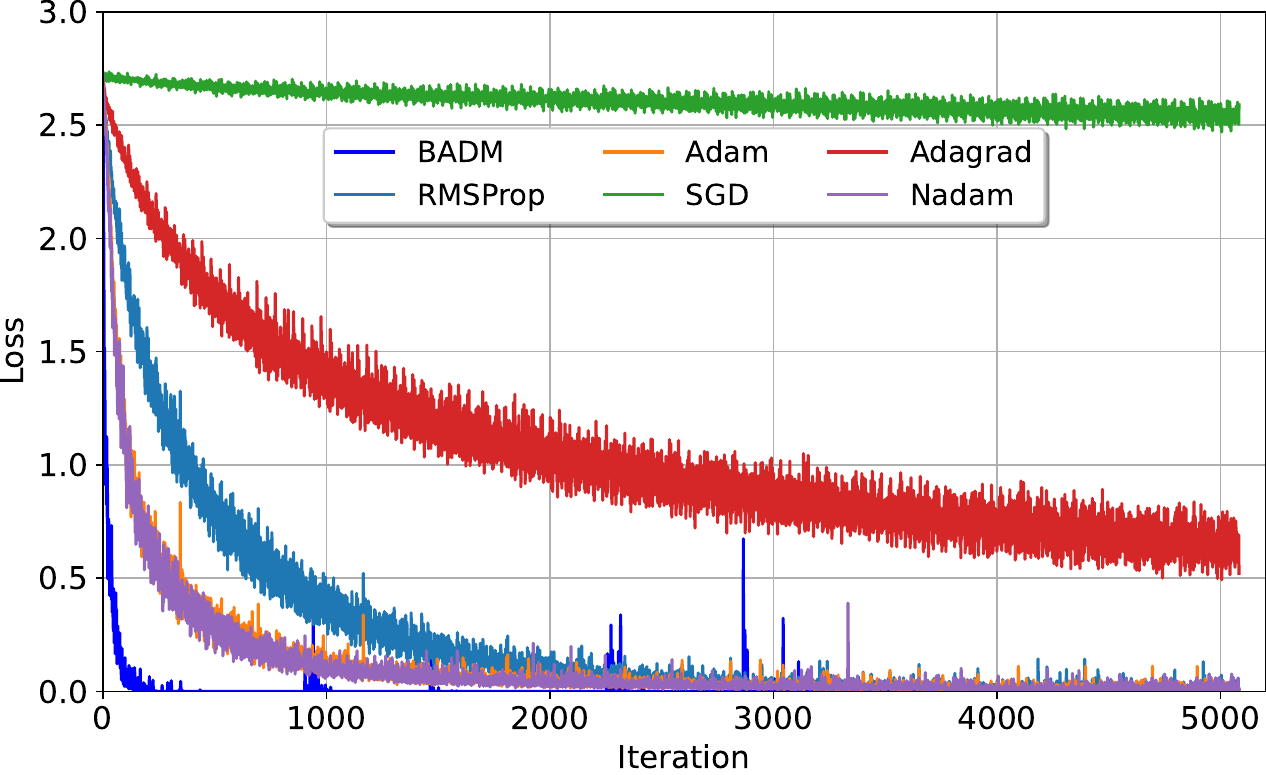} 
	\caption{GNN for CS}
\end{subfigure} 
\begin{subfigure}{.325\textwidth}
	\centering
	\includegraphics[width=.98\linewidth]{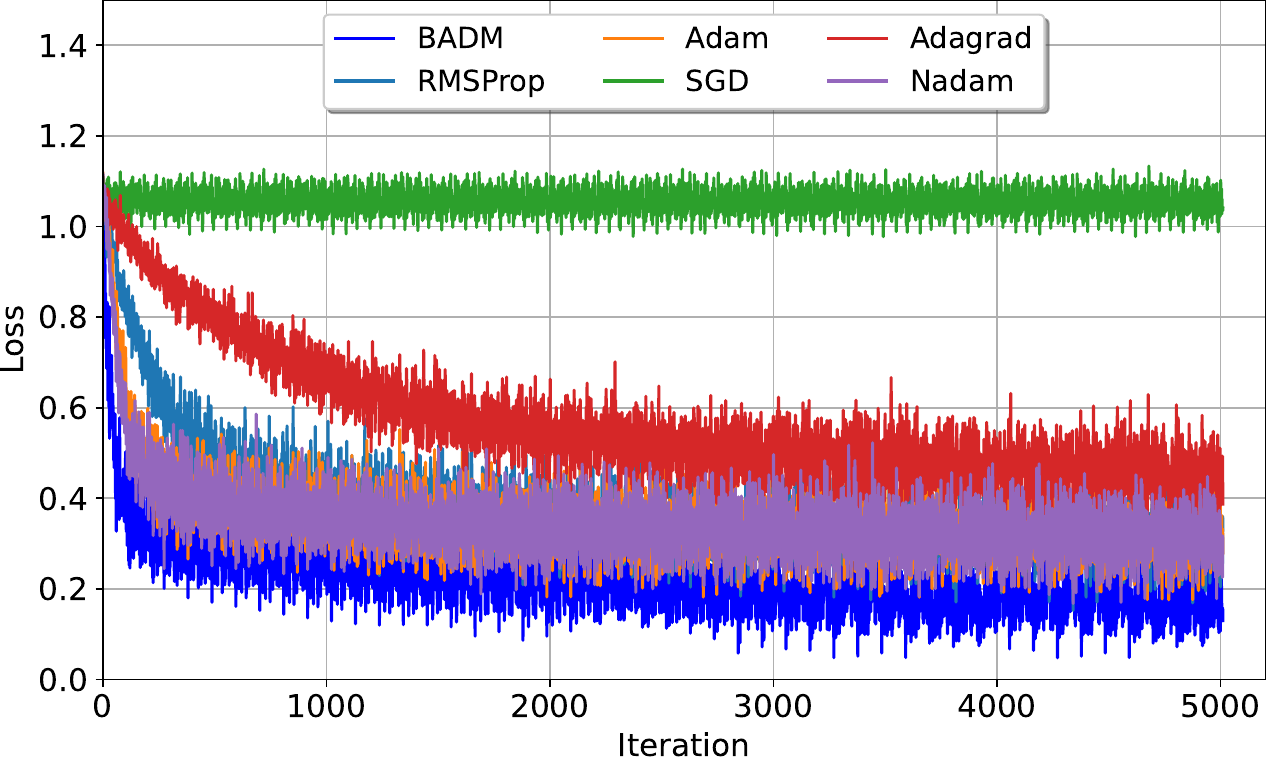} 
	\caption{GNN for Pubmed}
\end{subfigure}
\caption{Training loss v.s. iterations for node classification. \label{fig:DNN_node_classification}} \vspace{5mm}
\end{figure*} 
    
 As reported in Table \ref{table:1},   BADM attains the highest testing accuracy for most cases among all optimizers. Moreover, from Fig. \ref{fig:DNN_node_classification}, BADM has the fastest convergence speed. Although the GNN trained by BADM doesn't achieve the best testing accuracy when classifying dataset 'Computers' (resp. `Physics'), BADM only requires 2300 (resp. 700) training iterations to reach that accuracy, while the others need at least 6800 (resp. 1380) iterations to achieve the same accuracy.      

One can observe that Adam and RmsProp, one of the most commonly used tools in DL, perform better than or equal to AdaGrad, SGD, NAdam, and dlADMM. Therefore,  we will only select them as benchmarks for the remaining tasks in the sequel.

\textbf{b) Graph properties prediction.} 
Molecular structures can be naturally represented as an undirected graph, and  GNNs (e.g., MPNN) have proven effective for predicting molecular properties. The dataset used for this experiment consists of 2,050 molecules, each identified by a name, label, and SMILES string. The blood-brain barrier (BBB) is a membrane that separates the blood from the brain's extracellular fluid, preventing most drugs from entering the brain. Studying this barrier is essential for developing new drugs targeting the central nervous system. The dataset labels are binary (1 or 0), indicating whether the molecules can permeate the BBB.

\begin{figure*}[!t]
\centering
\begin{subfigure}{.325\textwidth}
	\centering
	\includegraphics[width=.98\linewidth]{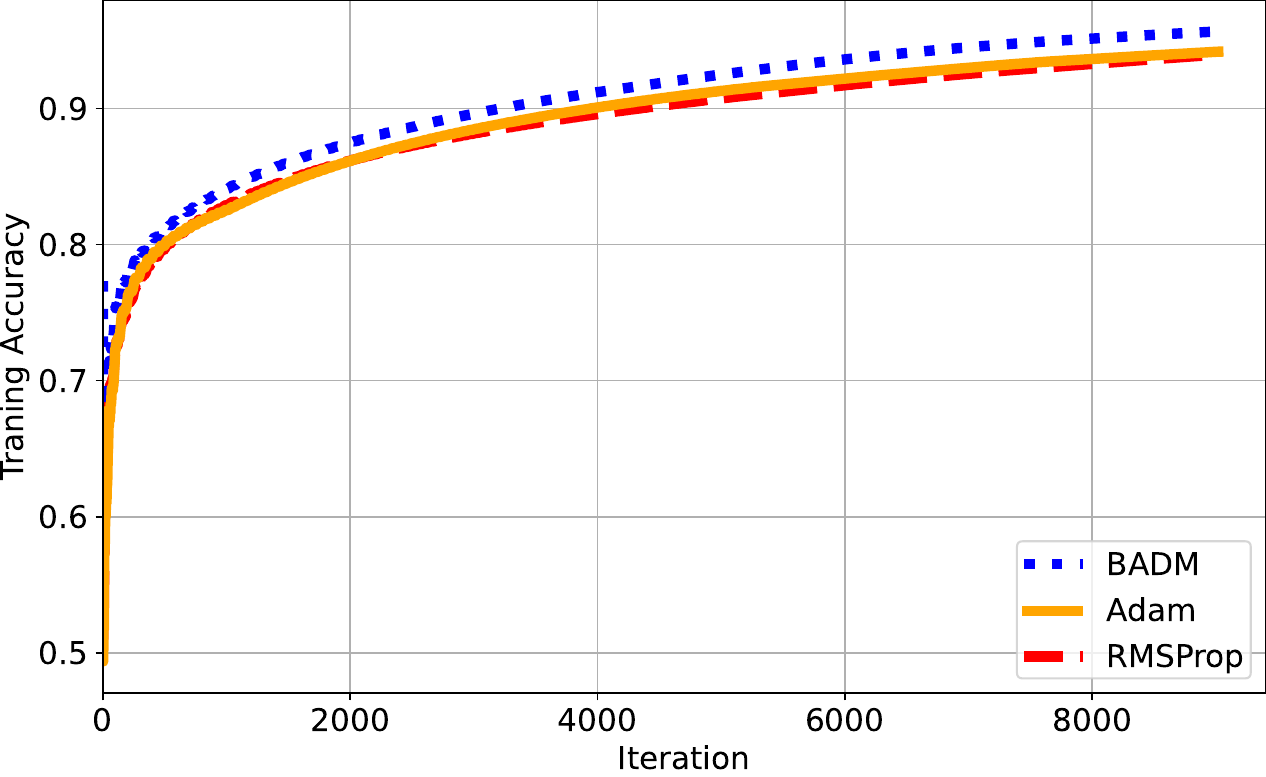}
    \caption{Training acc. for graph prediction}
    \label{fig:Train_molecular}
\end{subfigure}	  
\begin{subfigure}{.325\textwidth}
	\centering
	\includegraphics[width=.98\linewidth]{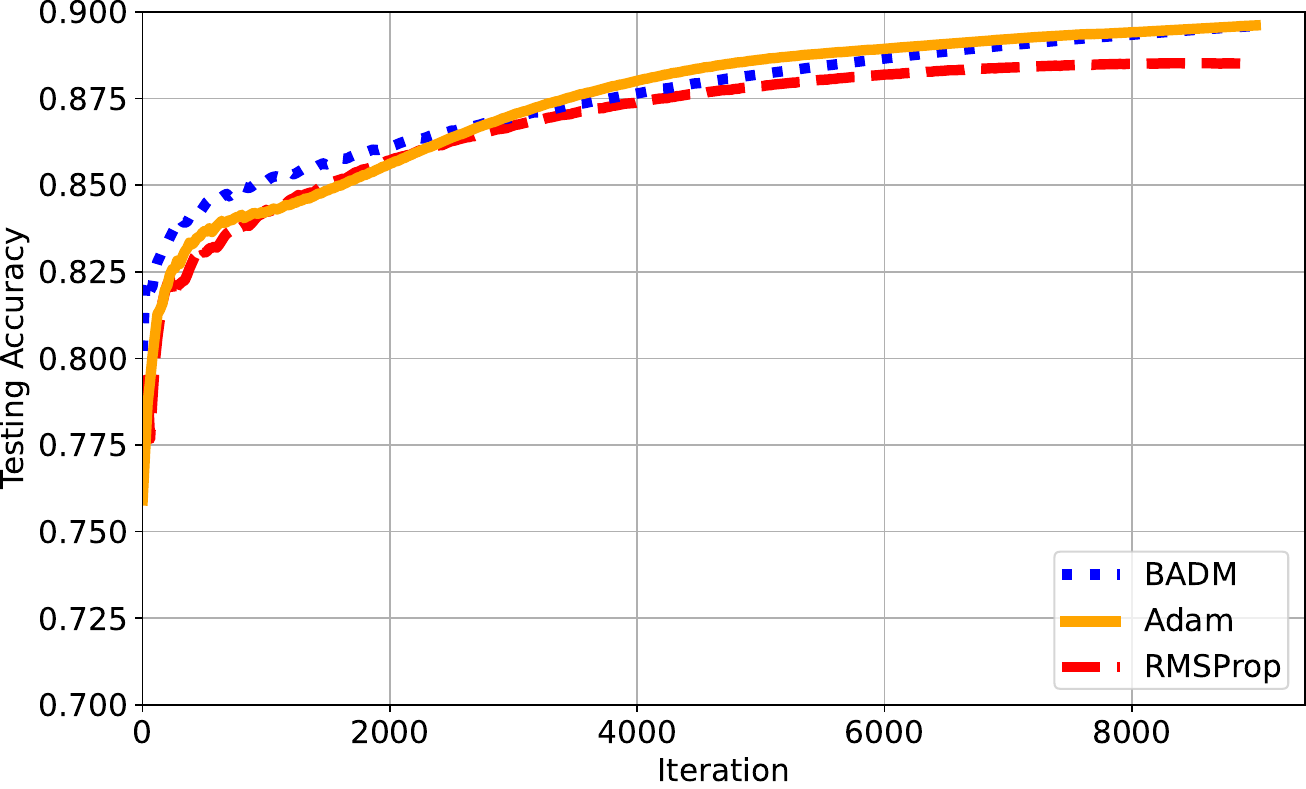}
    \caption{Testing acc. for graph prediction}
    \label{fig:Test_molecular}
\end{subfigure}
\begin{subfigure}{.325\textwidth}
	\centering
	\includegraphics[width=.98\linewidth]{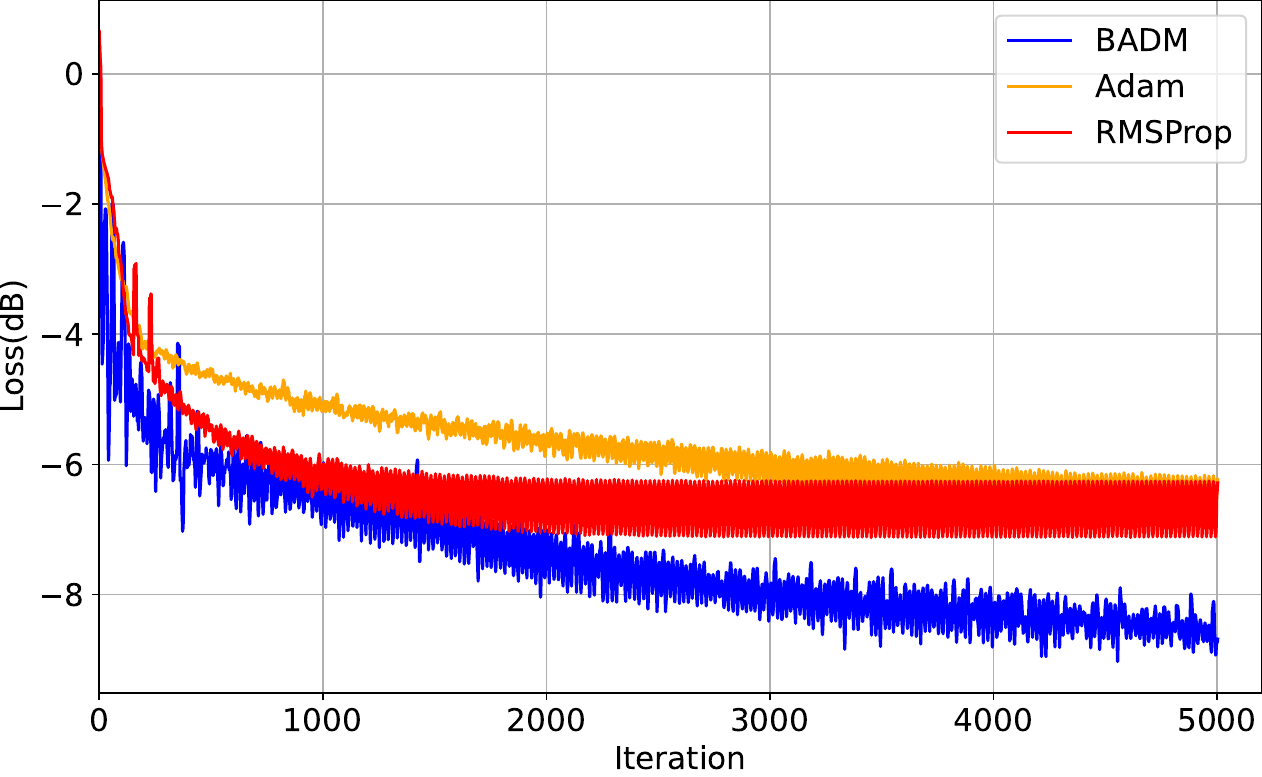} 
    \caption{Training loss for object detection}
    \label{fig:object_detection}
\end{subfigure}
\caption{Performance of three optimizers. 
 \label{fig:molecular}}
\end{figure*}

Following the approach introduced by \cite{gilmer2017neural}, we implement MPNN with three stages: message passing, readout, and classification. In the first stage, the edge network passes messages from the 1-hop neighbours of a node to another using their edge features, which updates the node features. Then a recurrent neural network updates the most recent node state using previous node states, allowing information to transfer from one node to another. The second stage (i.e., the readout) converts the $k$-step-aggregated node states into graph-level embeddings for each molecule. The last stage employs a two-layer classification network to predict BBBP. Figs. \ref{fig:Train_molecular} and \ref{fig:Test_molecular} present the training and test accuracy against the iteration. BADM reaches the best training accuracy and achieves the same testing accuracy as that generated by Adam, which is better than RMSProp.

\subsection{Computer vision}
This section focuses on Computer vision tasks, such as image classification and detection. For the classification task, we implement a CNN with various images with sizes ranging from $32\times32$ to $256\times256$  and classes ranging from $3$ to $15$. For the detection task, we implement a vision transformer (ViT)  \cite{dosovitskiy2020image} to train the Caltech 101 dataset to detect an airplane in the given image.

\textbf{c) Image classification.}  
Six datasets used for this experiment are `Cifar-10', `Svhn', `Deep weeds', `Cmaterdb', `Patch Camelyon', and `TF Flowers'.
'Cifar-10' includes 60,000 color images of size ${32 \times 32 \times 3}$, divided into 10 classes with 6,000 images per class. It provides 50,000 images for training and 10,000 images for testing.  
'Svhn' is designed for digit recognition and contains over 600,000 real-world images, each sized ${32 \times 32 \times 3}$ and categorized into 10 classes. 
'Deep weeds' has 17,509 images of size ${256 \times 256 \times 3}$, capturing $8$ different weed species native to Australia along with neighbouring flora. 
`Cmaterdb' includes handwritten numerals in Bangla, Devanagari, and Telugu. Images are ${32 \times 32 \times 3}$ RGB-colored and divided into ten classes.
`Patch Camelyon' comprises 327,680 color images of size ${96 \times 96 \times 3}$ extracted from histopathologic scans of lymph node sections, with each image annotated with a binary label indicating the presence of metastatic tissue. 
`TF Flowers' contains images of daisies, dandelions, roses, sunflowers, and tulips, each sized ${180 \times 180 \times 3}$.

\begin{table}[h]
\renewcommand{\arraystretch}{1.5}\addtolength{\tabcolsep}{-2.8pt}
    \centering
    \caption{Hyperparameters and testing accuracy for image classification.}
    \label{table:2}
    \begin{tabular}{ccccccc}
        \hline
        & Cifar-10 & Svhn   & Weeds & Cmaterdb & Camelyon & Flowers\\
        \hline
        $B$& 256 & 128   & 64& 128 & 128 & 64\\  
        $S$& 64 & 32  & 16& 32 & 32 & 16 \\  
       $\rho$ & 8500  & 7000 & 8000& 5000 & 7000 & 7000\\ 
       $\sigma$ & 1500    & 3000 & 2000& 5000 & 3000 & 3000\\ 
       \hline
       RMSProp& 0.7133 & 0.8946   & 0.6069 & 0.9739 & 0.8425 & 0.8437\\ 
       Adam & 0.7043 & 0.8932   & 0.6171 & 0.9740 & 0.8216 & 0.8593 \\ 
       BADM & \textbf{0.7149} & \textbf{0.9043}  & \textbf{0.6663} & \textbf{0.9750} & \textbf{0.8739} & \textbf{0.9218} \\ \hline

    \end{tabular}%
    
    \medbreak
    \end{table}

\begin{figure*}[!th]
\centering
\begin{subfigure}{.325 \textwidth}
	\centering
	\includegraphics[width=.98\linewidth]{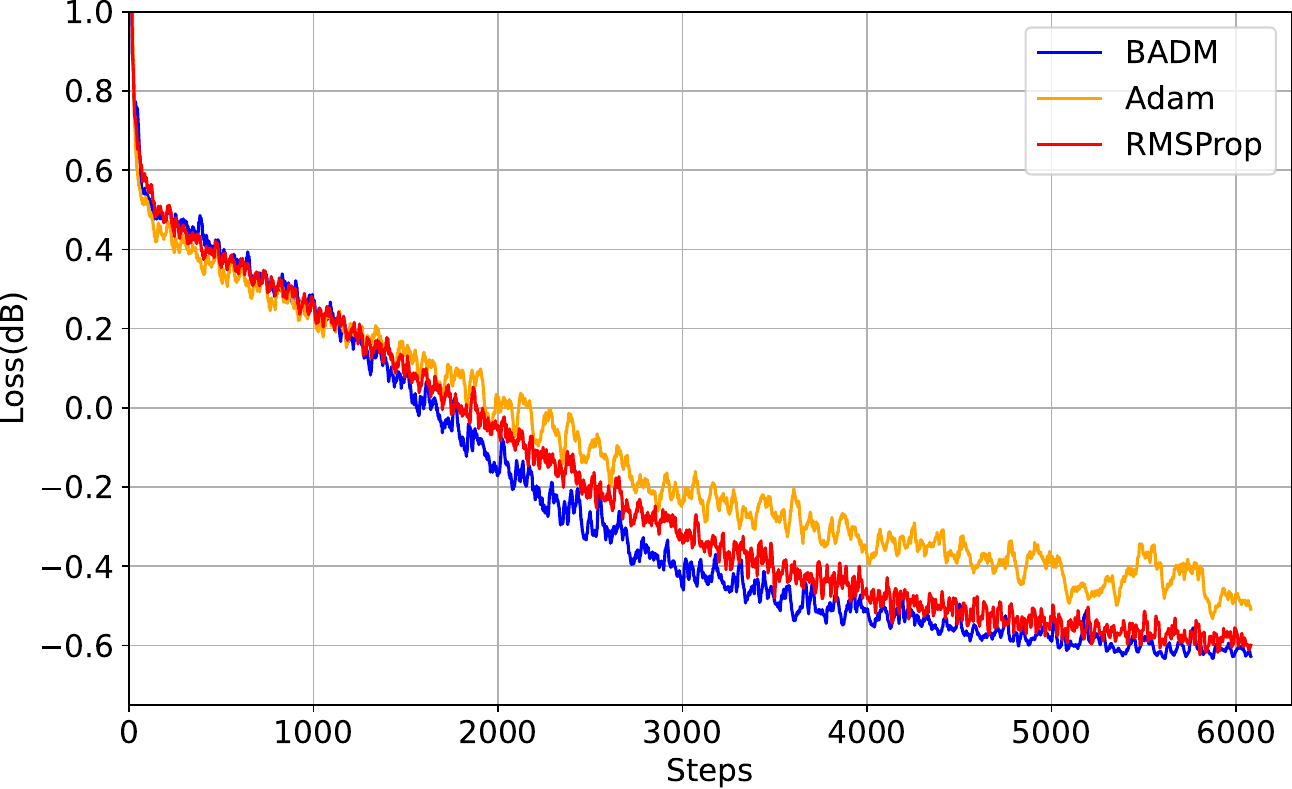} 
	\caption{Weeds}
\end{subfigure}	  
\begin{subfigure}{.325 \textwidth}
	\centering
	\includegraphics[width=.98\linewidth]{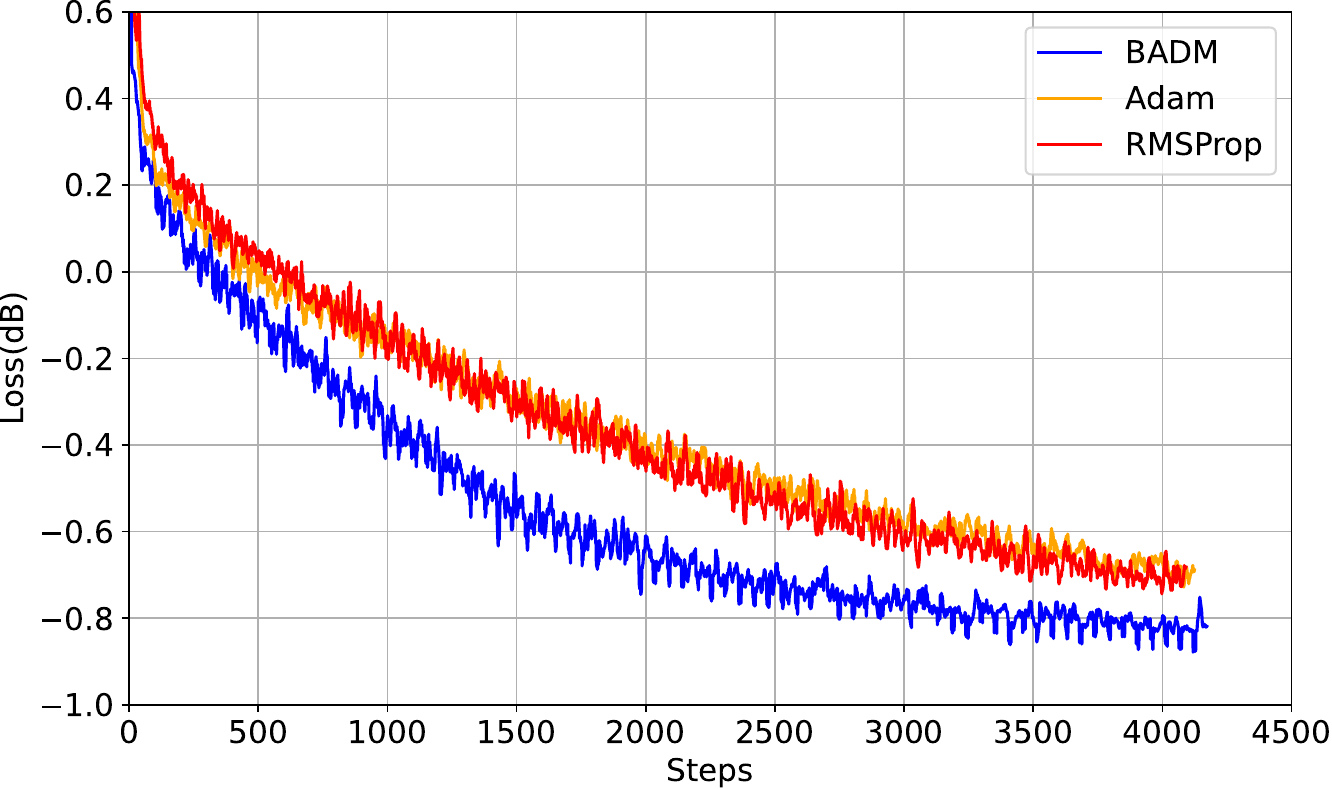} 
	\caption{Flowers}
\end{subfigure}	 
\begin{subfigure}{.325 \textwidth}
	\centering
	\includegraphics[width=.98\linewidth]{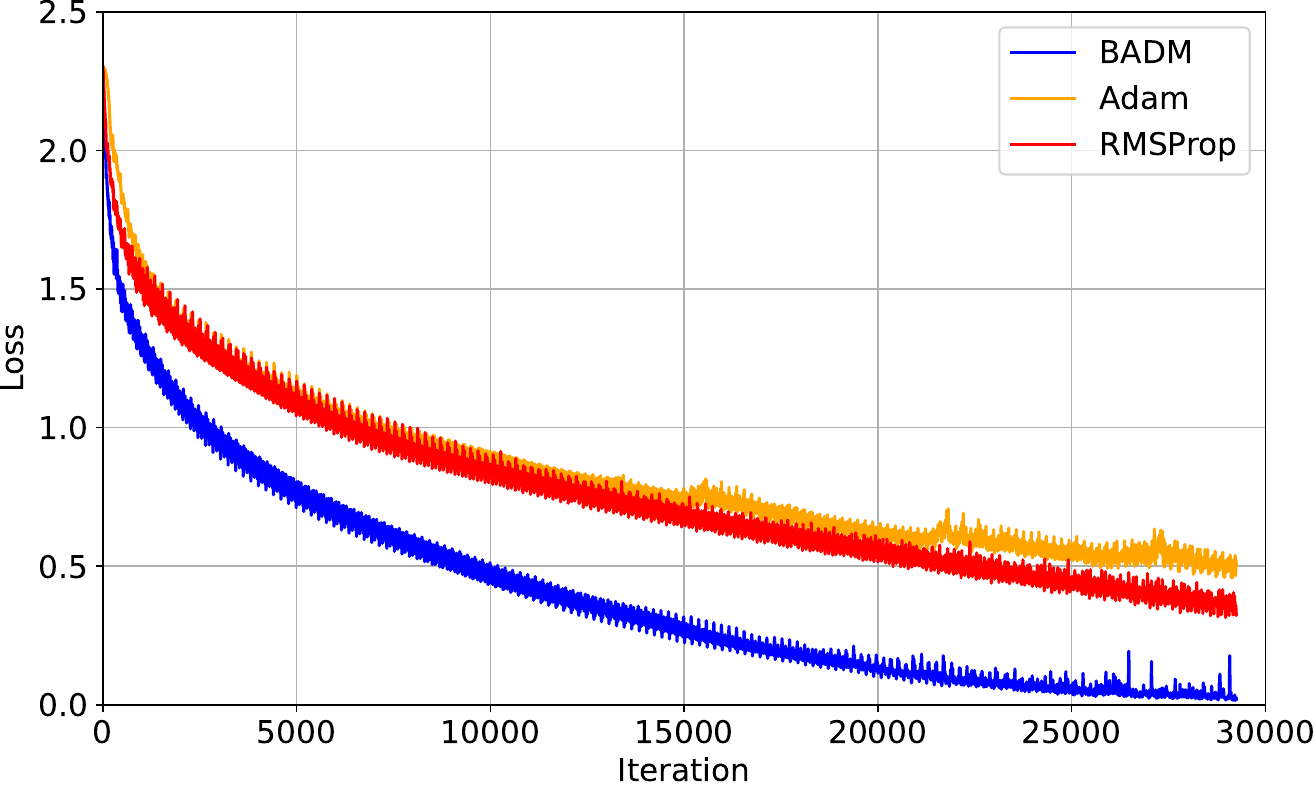} 
	\caption{Cifar-10}
\end{subfigure}  \\  \vspace{3mm}
\begin{subfigure}{.325 \textwidth}
	\centering
	\includegraphics[width=.98\linewidth]{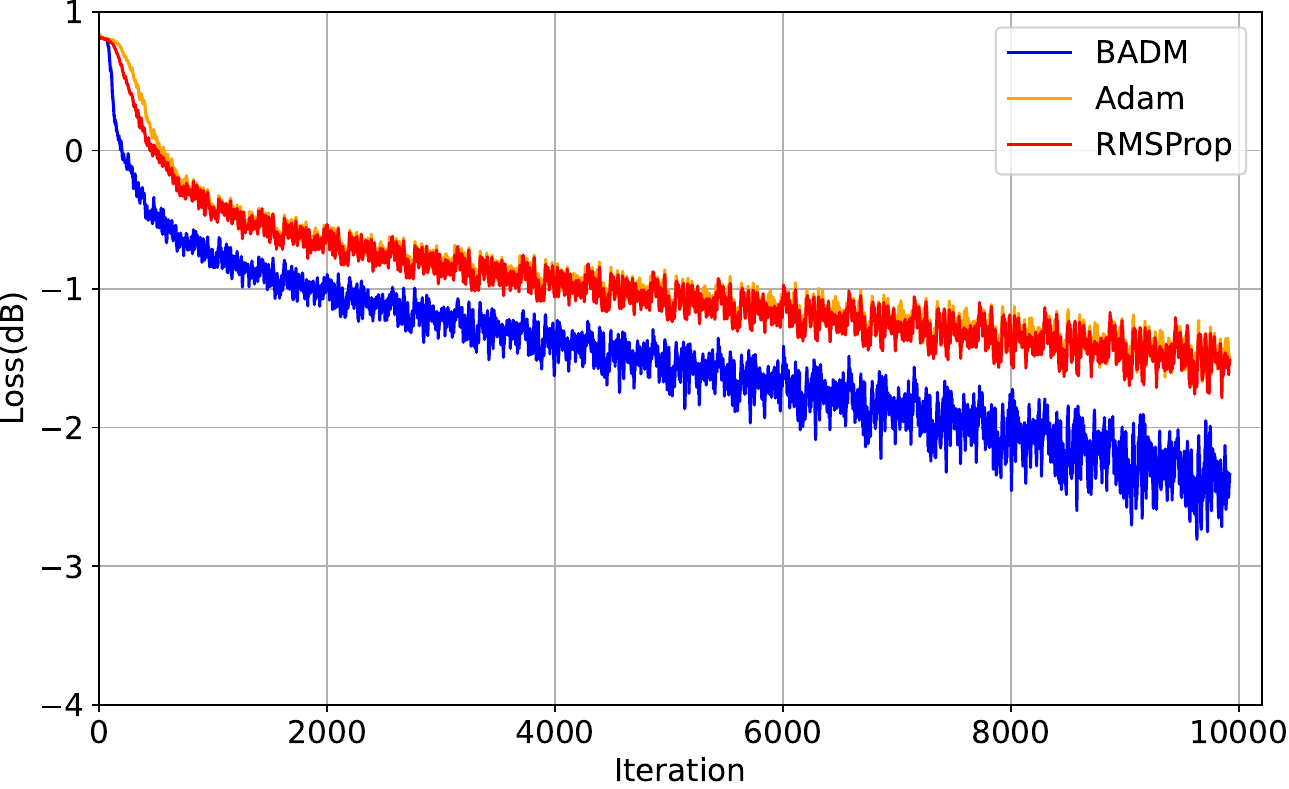} 
	\caption{Svhn}
\end{subfigure}
\begin{subfigure}{.325 \textwidth}
	\centering
	\includegraphics[width=.98\linewidth]{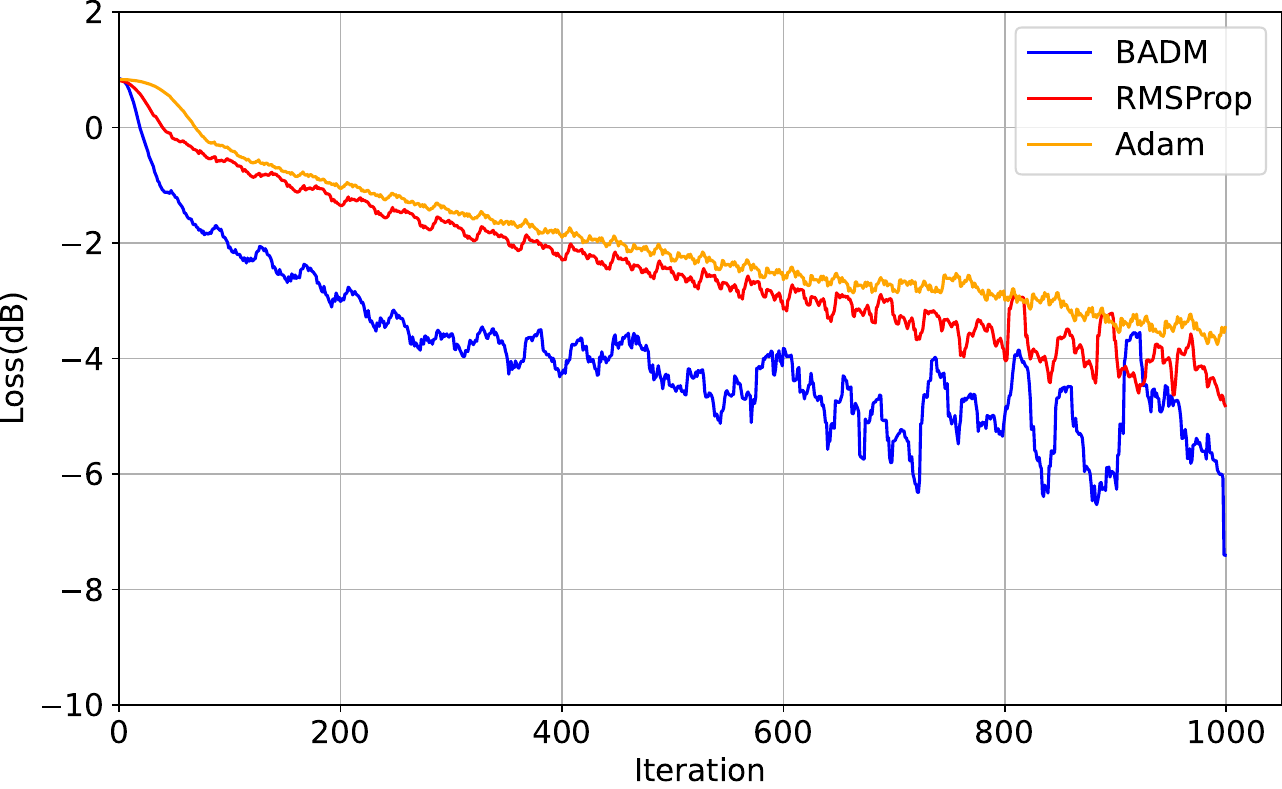} 
	\caption{Cmaterdb}
\end{subfigure} 
\begin{subfigure}{.325 \textwidth}
	\centering
	\includegraphics[width=.98\linewidth]{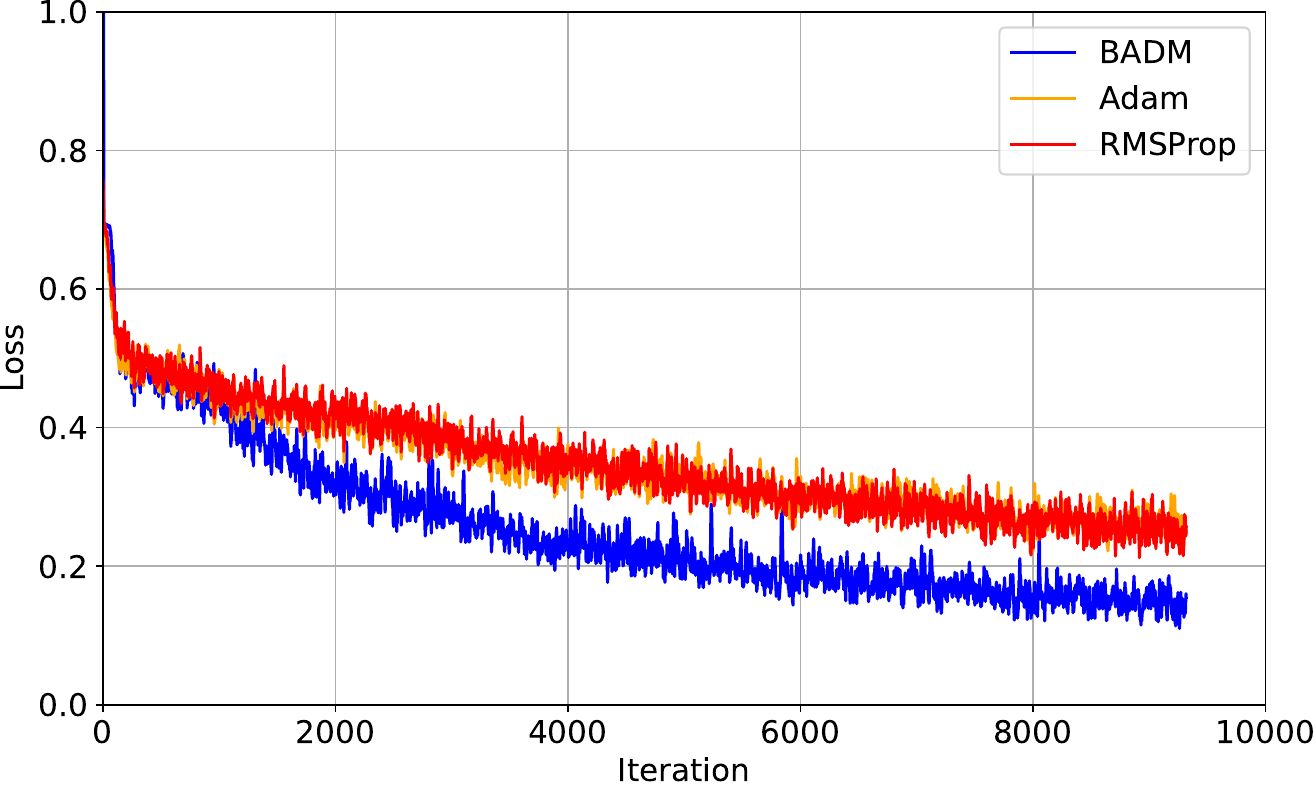} 
	\caption{Camelyon}
\end{subfigure}\\  
\caption{Training loss v.s. iterations for image classifications.\label{fig:image_classification}}
\end{figure*}

The architecture of the CNN consists of several 2D convolutional layers for feature extraction, each with $3\times 3$ kernel size, ReLU activation function, and max-pooling following each output. A fully connected layer with 64 neurons is employed before the final classification layer. The structure of convolutional layers and hyperparameters of experiments are reported in Table \ref{table:2}, where testing accuracy is also recorded. Compared to Adam and RMSProp, BADM attains the highest accuracy for all tasks. One can observe that the accuracy exhibits significant improvement for several challenging datasets, with enhancements of $4.92\%$ and $6.25\%$ observed for `Weeds' and `Flowers', respectively. The training loss along with the iterations for each algorithm is shown in Fig.  \ref{fig:image_classification}. Once again, BADM always attains the lowest training loss at each iteration. 

\textbf{d) Object detection.}  As demonstrated in \cite{dosovitskiy2020image},  a ViT was employed for object detection by predicting bounded box coordinates. Similarly, we train the ViT on the Caltech 101 dataset to detect an airplane in the image. Intersection over union (IOU) serves as a metric to assess the overlap between predicted and true bounded boxes.  All optimizers are trained on $640$ images with size ${224\times224}$. Fig.  \ref{fig:object_detection} shows that BADM has a faster convergence with the training steps rising. The average testing IOU for BADM, Adam, and RMSProp is 0.9214, 0.9120, and 0.9116, respectively. 

\begin{figure*}[!th]
\centering
\begin{subfigure}{.325 \textwidth}
	\centering
	\includegraphics[width=.95\linewidth]{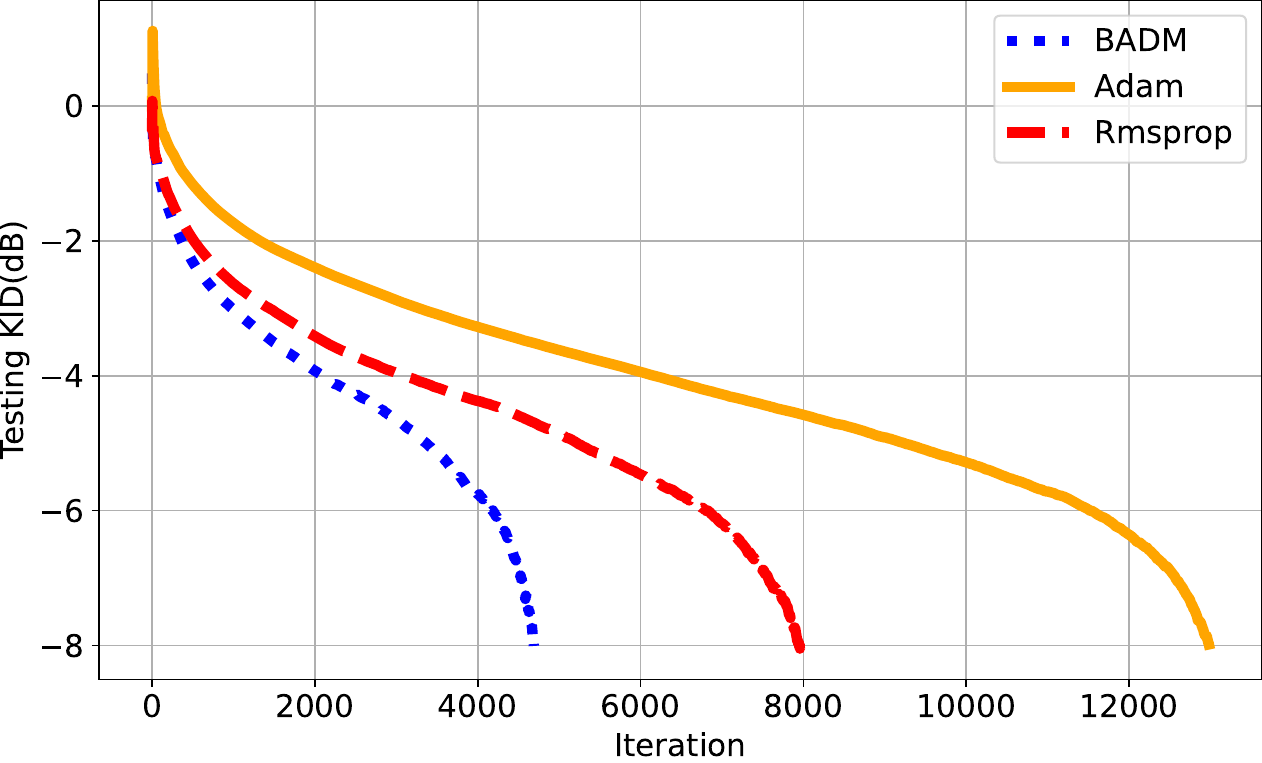} 
	\caption{MNIST}
	\label{fig:mnist-KID}
\end{subfigure}	 
\begin{subfigure}{.325 \textwidth}
	\centering
	\includegraphics[width=.95\linewidth]{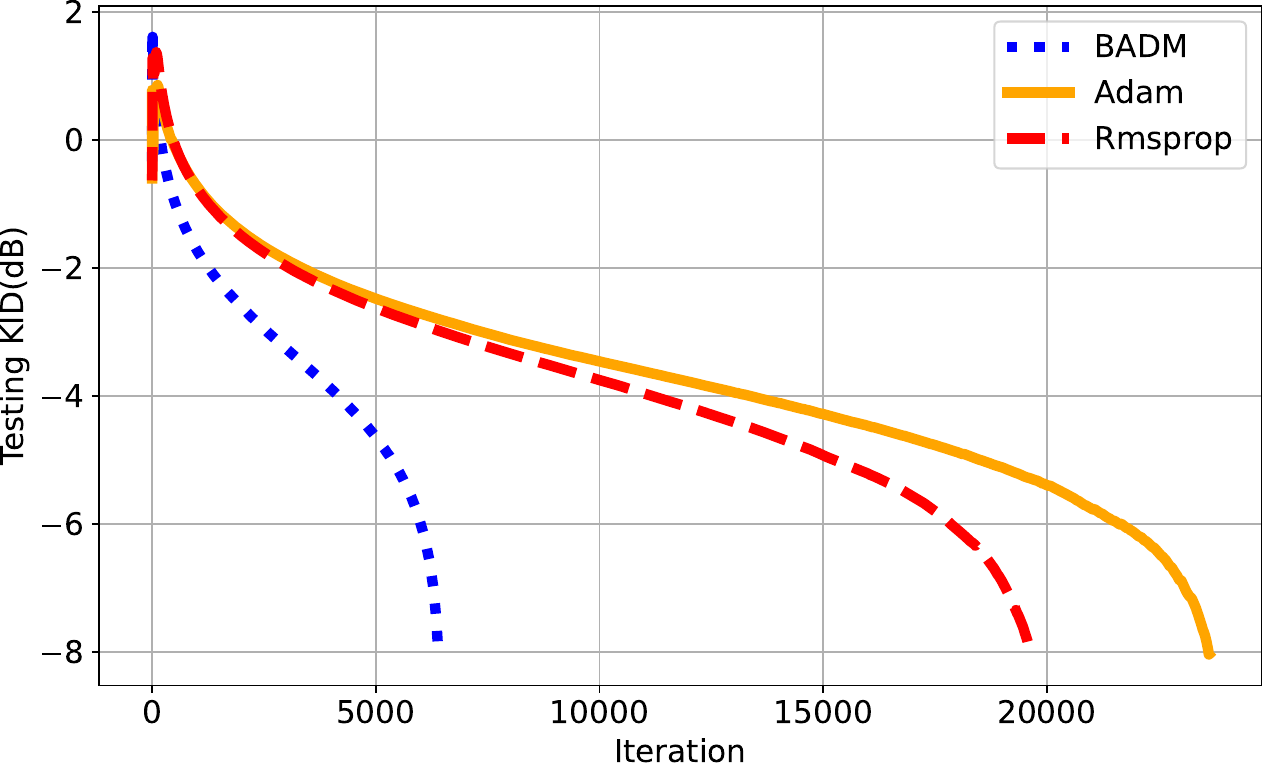} 
	\caption{Fashion-MNIST}
	\label{fig:fashion-KID}
\end{subfigure}   
\begin{subfigure}{.325 \textwidth}
	\centering
	\includegraphics[width=.98\linewidth]{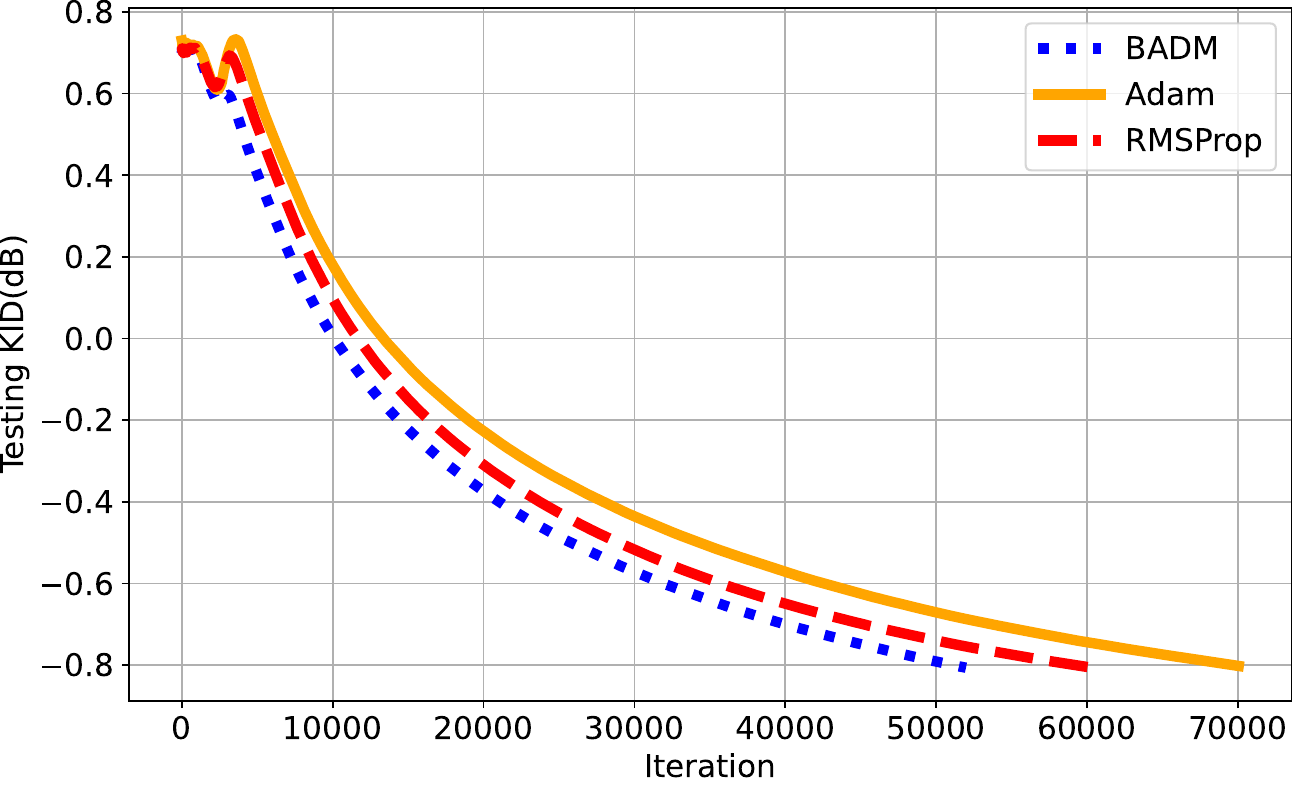} 
	\caption{DDIM.}
	\label{fig:oxford}
\end{subfigure}   
\caption{Testing KID v.s. iterations by conditional GAN and DDIM.\label{fig:Imagen}}
\end{figure*}

\subsection{Image generation}\label{sec:image-generation}
Generative models have gained significant attention due to their proficiency in synthesizing realistic images and have shown remarkable success in image generation lately \cite{goodfellow2020generative,ho2020denoising}. In this task, we examine the performance of BADM in both conditional and unconditional image synthesis. The evaluation metric is the kernel inception distance (KID), which computes the difference between generated and training distributions within the representation space of a pre-trained InceptionV3 network on ImageNet. A smaller KID value indicates higher similarity, reflecting the superior performance of the algorithm. KID is suitable for small-scale datasets because its expected value does not depend on the number of samples it is measured on.

\textbf{e) Conditional generative adversarial networks (GANs).} 
Compared with conventional GANs, conditional GANs allow us to control the appearance (e.g. class) of the generated samples. In an unconditional GAN, we begin by sampling noise of a fixed dimension from a normal distribution. Moreover, we incorporate class labels to the input channels for both generator (noise input) and discriminator (generated image input). In our experiments, We employ the conditional GAN framework introduced by \cite{mirza2014conditional} to perform $32\times32$ image generation on MNIST and Fashion-MNIST datasets, conditioned on digit classes. The discriminator is a CNN with 2 convolutional layers and one fully connected layer to classify whether the input image is generated or real. The generator consists of one fully connected layer following three convolutional layers to generate images. We employ two datasets,  `MNIST' and `fashion-MNIST', to train the conditional GAN. `MNIST' is a dataset containing 70,000 grayscale images of handwritten digits, each sized $28\times28$ pixels, categorized into 10 classes (digits 0 to 9). `Fashion-MNIST' consists of 70,000 grayscale images of fashion items, also $28\times28$ pixels, divided into 10 classes, including objects like shirts, trousers, and shoes.

For these experiments, the batch size, and sub-batch size are {$B=64$} and ${S=16}$ for all {$b\in\BB$}, the learning rate is 0.0001 for Adam and RMSProp,  and {$(\rho,\sigma)=(6000, 4000)$} for BADM.  As depicted in Figs.  \ref{fig:mnist-KID} and \ref{fig:mnist-KID}, BADM exhibits a much faster training speed to achieve the same KID score compared to RMSProp and Adam. For instance, when the KID score reaches $-8$ dB, the training speed of BADM is at least two times (resp. three times) faster than the others in generating MNIST (resp. Fashion-MNIST) images.

\begin{figure*}[!t]
\centering	 
\begin{subfigure}{.325 \textwidth}
	\centering
	\includegraphics[width=.99\linewidth]{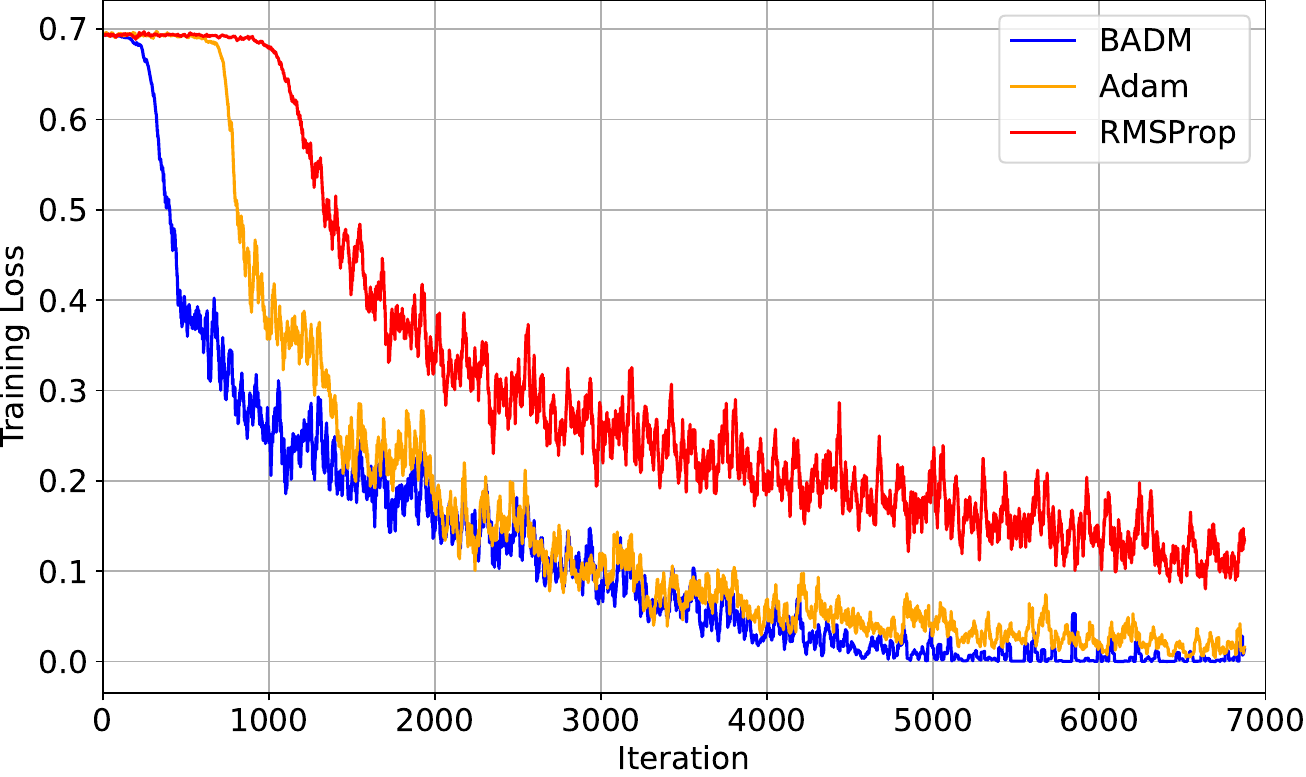} 
	\caption{Text classification}
	\label{fig:textclass}
\end{subfigure} 
\begin{subfigure}{.325 \textwidth}
	\centering
	\includegraphics[width=.95\linewidth]{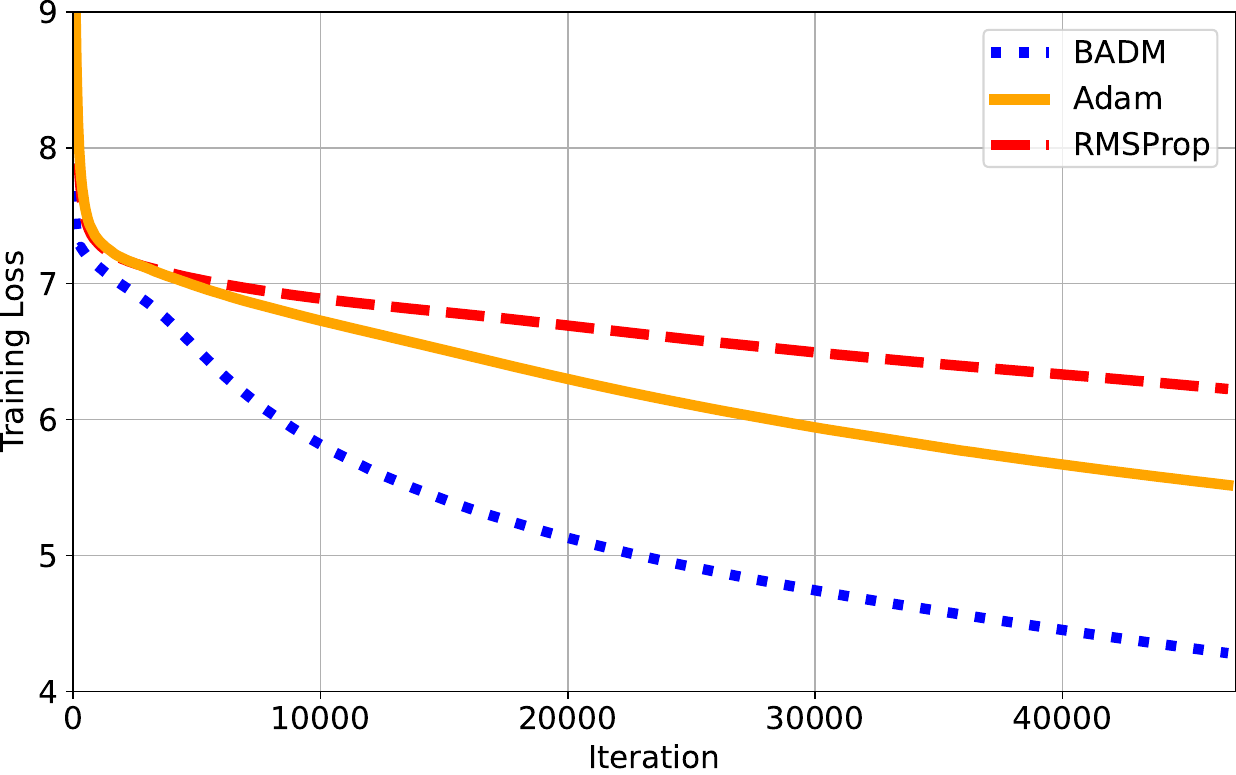} 
	\caption{Pre-training for MLM}
	\label{fig:mlmpre}
\end{subfigure} 
\begin{subfigure}{.325 \textwidth}
	\centering
	\includegraphics[width=.97\linewidth]{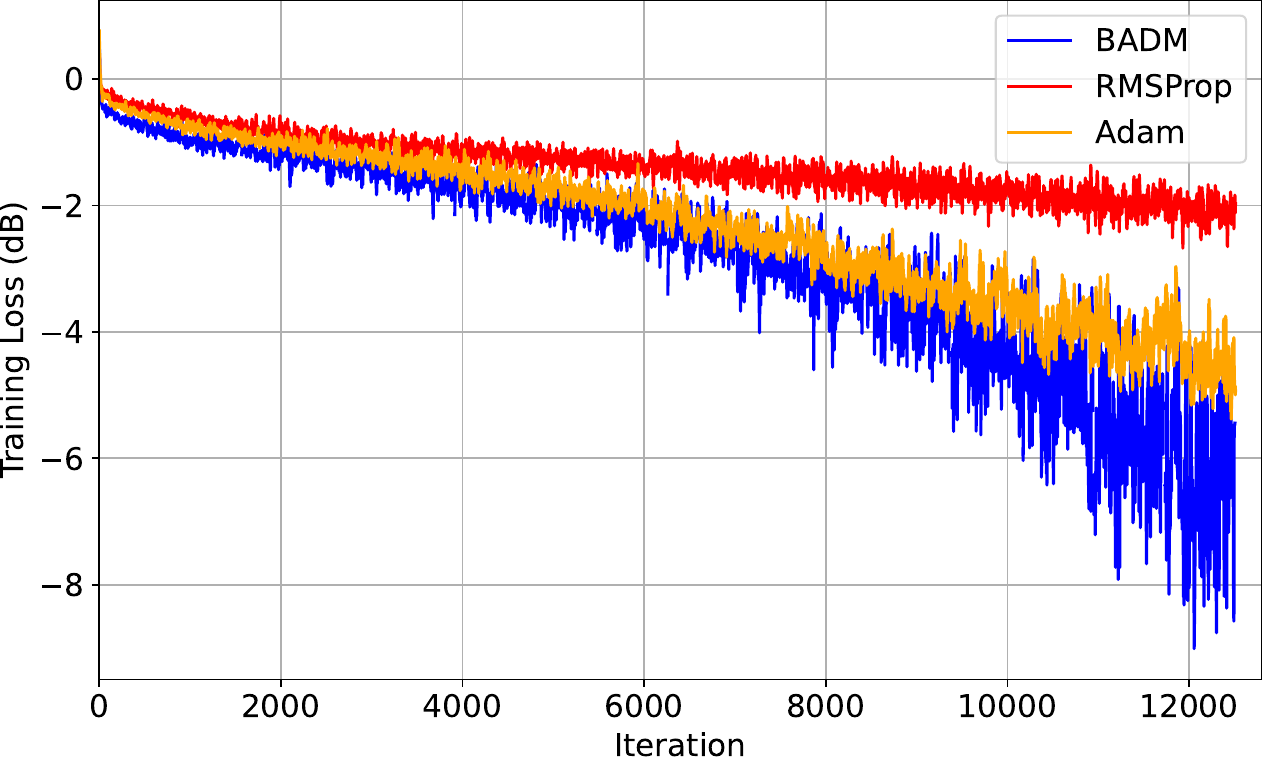} 
	\caption{Fine-tuning for MLM}
	\label{fig:mlmfine}
\end{subfigure}	 \\  
\caption{Training loss v.s. iterations for text classification and MLM.\label{fig:mlm}}
\end{figure*}
\textbf{f) Denoising diffusion implicit model (DDIM).} We implement another popular generative model, DDIM \cite{song2020denoising}, which can rival GANs in image synthesis quality but incurs higher training costs due to multiple forward passes needed for generating an image.
DDIM is a process that gradually transforms an image into noise. By simulating this process, we can create noisy versions of our training images and then train a neural network to denoise them. Once the network is trained, it can perform reverse diffusion during inference, allowing us to generate an image from noise. We use a U-Net with matching input and output dimensions as the architecture of the neural network for denoising. The U-Net network takes two inputs: the noisy images and the variances of their noise components. The variances are necessary because different noise levels necessitate different denoising operations. We transform these noise variances using sinusoidal embeddings, similar to positional encodings used in transformers. This operation enhances the network's sensitivity to noise levels, which is essential for optimal performance. The training process for DDIM involves uniformly sampling random diffusion times and mixing the training images with Gaussian noise at rates corresponding to the diffusion times. The model is then trained to separate the noisy image into its original image and noise components. Typically, the neural network is trained to predict the unscaled noise component, from which the image component can be derived using the respective signal and noise rates. In this context, the diffusion time is defined as the discrete steps in the forward diffusion process where noise is incrementally added to the data. 

The dataset is the Oxford Flowers 102, a diverse collection of approximately 8,000 images of various flower species and $20\%$ images are sampled dataset for real-time evaluation. DDIM is trained using this dataset with a batch size of 64. Since DDIM generates colourful images instead of grey-scale images in conditional GANs, this task is more complex, resulting in higher KID values. As illustrated in Fig.  \ref{fig:oxford}, BADM takes the least steps to derive the same KID.

\subsection{Natural Language Processing}
This subsection focuses on two NLP tasks. The numerical results demonstrate that BADM is capable of delivering a faster convergence while maintaining the same testing accuracy. Such improvement becomes more evident during pre-training of the language model. 

\textbf{g) Text classification from scratch.} Following the experimental setup in  \cite{sachan2019revisiting}, we demonstrate the workflow on the IMDB sentiment classification dataset. The IMDB dataset consists of 25,000 movie reviews with binary sentiment labels indicating positive or negative feedback. We use 20,000 reviews for training and 5,000 reviews for testing. At the beginning of the model, we employ a text vectorization layer for word splitting and indexing. This layer vectorizes the text into integer token IDs, transforming a batch of strings into a dense representation (one sample = 1D array of float values encoding an unordered set of tokens). The 1D CNN consists of 2 convolutional and fully connected layers, each with 128 kernels. In the experiment, the batch and sub-batch sizes are 32 and 8, the learning rate is 0.0002 for Adam and RMSProp,  and $(\rho,\sigma)=(1000, 4000)$ for BADM. 

From Fig.  \ref{fig:textclass}, the training loss for BADM declines dramatically when the iteration number is between 400 and 1000. The testing accuracy for BADM and RMSProp are both 0.9375, which is higher than  0.9062 attained by Adam. Nevertheless, RMSProp displays a slower convergence and its training loss fluctuates significantly.

\textbf{h) End-to-end masked language modelling (MLM).} 
MLM is a fill-in-the-blank task, where a model uses the context words surrounding a mask token to predict what the masked word is. 

Based on the example in \cite{salazar2019masked}, the IMDB dataset is used again to evaluate the performance of three algorithms for this task. Similar to the task of text classification, a text vectorization layer is employed. Additionally, we apply a mask function to the input token IDs, randomly masking 15\% of all tokens in each sequence. We then construct a BERT-like pre-training model that includes a Multi-Head-Attention layer, which takes token IDs (including masked tokens) as inputs and predicts the correct IDs for these masked tokens.  The pre-training model consists of a text vectorization layer, a multi-head attention layer, two fully-connected layers to process the attention output, and one fully-connected layer to predict the masked tokens. After pre-training, we fine-tune a sentiment classification model by creating a classifier by adding a pooling layer and a dense layer on top of the pre-trained BERT features. 

For the pre-training experiment, the batch and sub-batch sizes are 32 and 8, the learning rate is 0.0001 for Adam and RMSProp,  and $(\rho,\sigma)=(8000, 2000)$ for BADM. For the fine-tuning experiment, all parameters remain the same except for $(\rho,\sigma)=(5000, 5000)$ for BADM. As shown in Fig.  \ref{fig:mlmpre}, BADM significantly improves convergence speed during pre-training. The fine-tuning performance is illustrated in Fig.  \ref{fig:mlmfine}, where all optimizers achieve the same test accuracy of 0.9062, and it can be clearly observed that the training loss of BADM declines the fastest.

\section{Conclusion} 
Based on the conventional DL training process, we developed an ADMM-type learning algorithm that differs from the conventional ADMM cast to train neural networks. The proposed algorithm can be deemed as a data-driven approach and flexible enough to be applied across a wide range of applications. Extensive numerical experiments have demonstrated its superior performance compared to other state-of-the-art solvers.


\bibliographystyle{IEEEtran}
\bibliography{BADMreference}

\begin{wrapfigure}{l}{25mm} 
\includegraphics[width=1in,height=1.25in,clip,keepaspectratio]{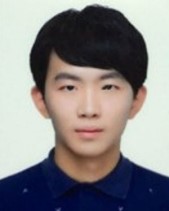}
\end{wrapfigure}\par
\textbf{Ouya (Tracy) Wang} received the B.Eng. degree in electrical and electronic engineering from the University of Manchester in 2020, and the M.Sc. degree in applied machine learning from Imperial College London in 2021, where he is currently pursuing the Ph.D. degree with the Department of Electrical and Electronic Engineering. His research interests include accretionary learning and deep learning with application in signal processing.\par
\vspace{10mm} 
\begin{wrapfigure}{l}{25mm} 
\includegraphics[width=1in,height=1.25in,clip,keepaspectratio]{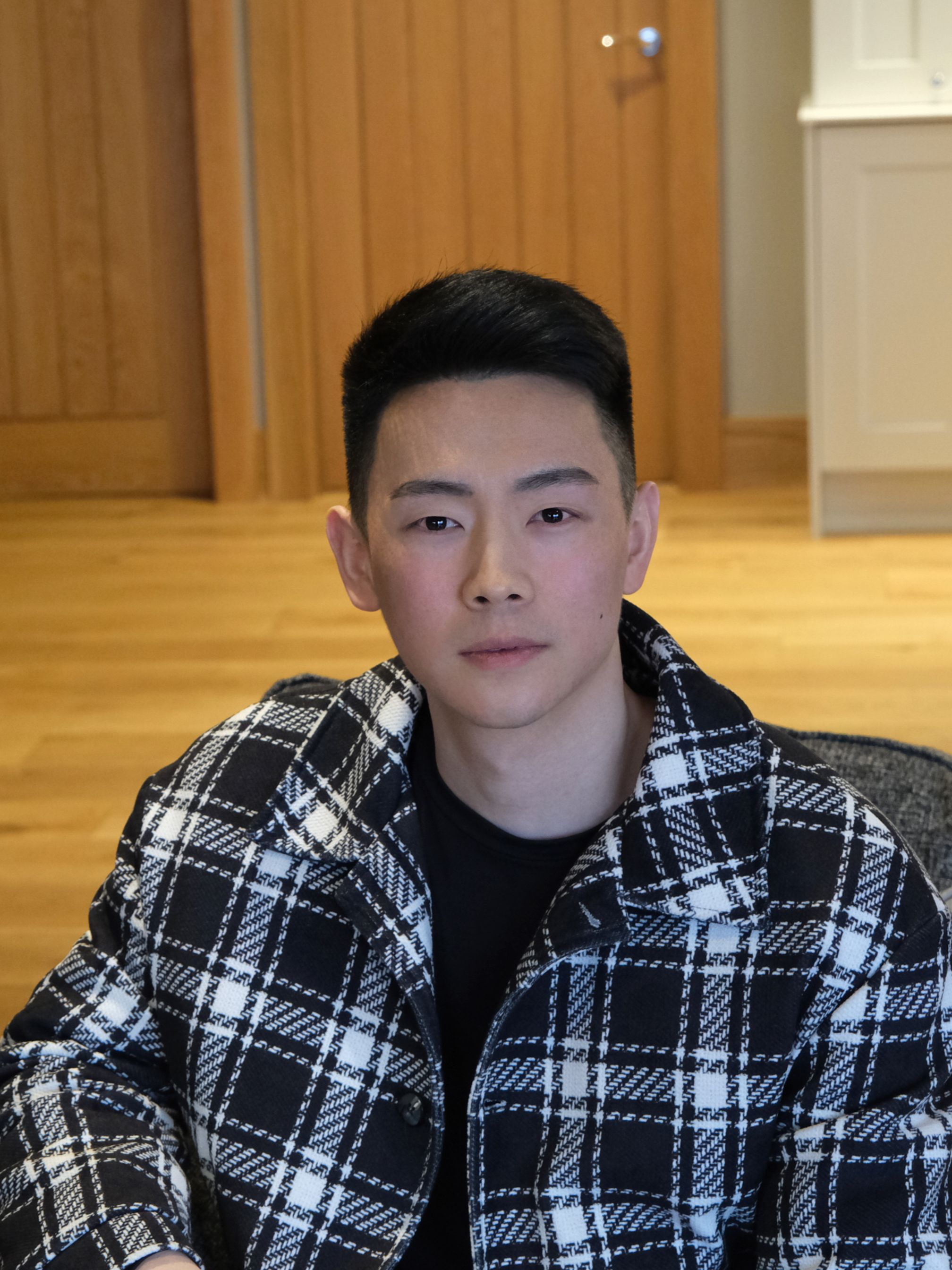}
\end{wrapfigure}\par
\textbf{Shenglong Zhou} is currently a Professor with Beijing Jiaotong University, China.  He received a Ph.D. degree in operational research from the University of Southampton, U.K., in 2018, where he was a Research Fellow and a Teaching Fellow. From 2021 to 2023, he was a Research Fellow with Imperial College London, U.K.  His research interests include the theory and methods for optimization in the areas of sparse optimization, 0/1 loss optimization, low-rank matrix optimization, bilevel optimization, and machine learning-related optimization.\par
\vspace{10mm} 
\begin{wrapfigure}{l}{25mm} 
\includegraphics[width=1in,height=1.25in,clip,keepaspectratio]{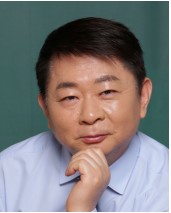}
\end{wrapfigure}\par
\textbf{Geoffrey Ye Li} (Fellow, IEEE) is currently a Chair Professor with Imperial College London, U.K. Before joining Imperial College London in 2020, he was a Professor with the Georgia Institute of Technology, USA, for 20 years and a Principal Technical Staff Member with AT\&T Labs—Research (previously, Bell Labs), Murray Hill, NJ, USA, for five years. He made fundamental contributions to orthogonal frequency-division multiplexing for wireless communications, established a framework on resource cooperation in wireless networks, and introduced deep learning to communications. In these areas, he has published over 600 journal and conference papers in addition to over 40 granted patents. His publications have been cited over 65 000 times with an H-index of 116. He has been listed as a Highly Cited Researcher by Clarivate/Web of Science almost every year. He won the 2024 IEEE Eric E. Sumner Award and several awards from IEEE Signal Processing, Vehicular Technology, and Communications Societies, including the 2019 IEEE ComSoc Edwin Howard Armstrong Achievement Award. He was elected to IEEE Fellow and IET Fellow for his contributions to signal processing for wireless communications.\par

\end{document}